\documentclass[11pt,a4paper]{article}
\usepackage{acl2018}
\usepackage{times}
\usepackage{latexsym}

\usepackage{url}

\usepackage{booktabs}
\usepackage{multirow}

\usepackage{microtype} % suppress hyphenations
\usepackage{outlines} % easy itemization
\usepackage{amsmath,amssymb}
\usepackage{mathtools}
\mathtoolsset{showonlyrefs=true} % equation number only if it's referred
\usepackage{bm} % bold math
\usepackage{caption}
\usepackage{subcaption}
\usepackage{xspace}
\usepackage{makecell} % line breaks inside a table cell

\makeatletter
\newcommand\footnoteref[1]{\protected@xdef\@thefnmark{\ref{#1}}\@footnotemark}
\makeatother

\aclfinalcopy % Uncomment this line for the final submission
\def\aclpaperid{1601} %  Enter the acl Paper ID here

\newcommand{\mat}[1]{\bm{#1}}
\renewcommand{\vec}[1]{\bm{#1}}
\newcommand{\setoftriples}{\mathcal{T}}
\newcommand{\setofents}{\mathcal{E}}
\newcommand{\setofrels}{\mathcal{R}}

\newcommand{\mtriple}[3]{\langle #1, #2, #3 \rangle}

\DeclareMathOperator*{\relu}{ReLU}
\DeclareMathOperator*{\tr}{tr}

\title{Interpretable and Compositional Relation Learning by Joint Training with an Autoencoder}

\author{Ryo Takahashi*\textsuperscript{1} \and
  Ran Tian*\textsuperscript{1} \and
  Kentaro Inui\textsuperscript{1,2}\\
  {\bf (* equal contribution)}\\
  \textsuperscript{1}Tohoku University\quad\textsuperscript{2} RIKEN, 
  \quad Japan \\
  {\tt \{ryo.t,\;tianran,\;inui\}@ecei.tohoku.ac.jp}
  \\}

\date{}

\begin{document}
\maketitle
\begin{abstract}
Embedding models for entities and relations are extremely 
useful for recovering missing facts in a knowledge base. Intuitively, a relation can be modeled by 
a matrix mapping entity vectors. However, relations reside 
on low dimension sub-manifolds in the parameter space of arbitrary matrices -- 
for one reason, composition of two relations $\mat{M}_1,\mat{M}_2$ may match a third $\mat{M}_3$
(e.g. composition of relations \texttt{currency\_of\_country} and \texttt{country\_of\_film} usually matches 
\texttt{currency\_of\_film\_budget}), which imposes compositional 
constraints to be satisfied by the parameters (i.e. $\mat{M}_1\cdot \mat{M}_2\approx \mat{M}_3$). 
In this paper we investigate a dimension reduction technique by training relations 
jointly with an autoencoder, which is expected to 
better capture compositional 
constraints. We achieve state-of-the-art on Knowledge Base Completion 
tasks with strongly improved Mean Rank, and show that joint training with 
an autoencoder leads to interpretable sparse codings of relations, helps discovering 
compositional constraints and benefits from compositional training. Our source code is released at
\url{github.com/tianran/glimvec}.
\end{abstract}

\section{Introduction}

Broad-coverage knowledge bases (KBs) such as %YAGO~\citep{DBLP:conf/www/SuchanekKW07} and 
Freebase~\citep{DBLP:conf/sigmod/BollackerEPST08} and 
DBPedia~\citep{auer2007dbpedia}
store a large amount of facts 
in the form of 
$\langle$\text{head entity}, \text{relation}, \text{tail entity}$\rangle$ triples 
(e.g. $\langle$\textit{The Matrix}, \texttt{country\_of\_film}, 
\textit{Australia}$\rangle$), which 
could support a wide range of reasoning and question answering applications. 
The Knowledge Base Completion (KBC) task aims to predict the missing part of an incomplete triple, 
such as $\langle$\textit{Finding Nemo}, \texttt{country\_of\_film}, ?$\rangle$, by reasoning 
from known facts stored in the KB.

\begin{figure}[!t]
\centering
\includegraphics[width=0.9\columnwidth]{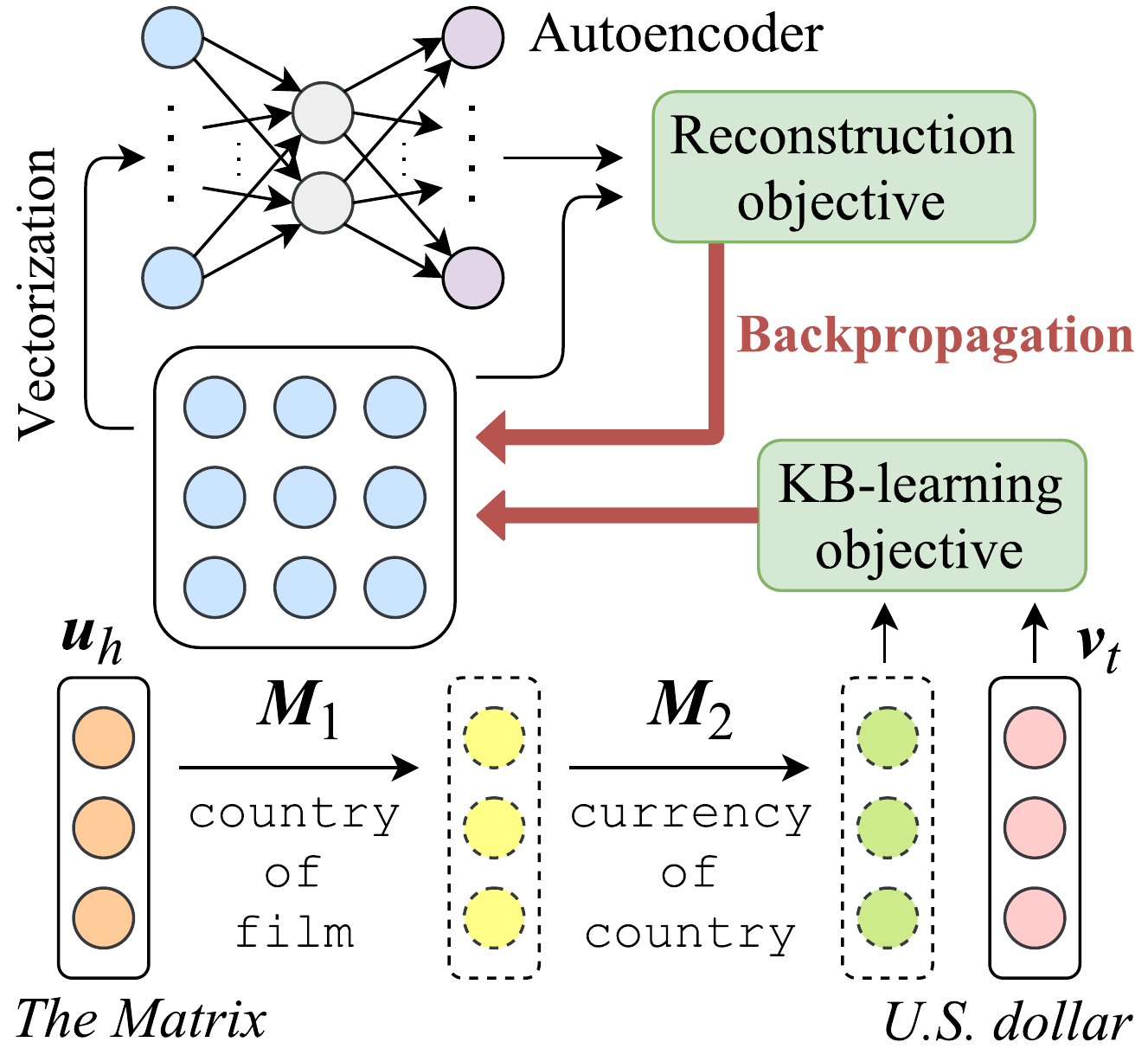}
\caption{In joint training, relation parameters (e.g. $\mat{M}_1$) 
receive updates from both
a \emph{KB-learning objective}, trying to predict entities in the KB; 
and 
a \emph{reconstruction objective} from an autoencoder, trying to 
recover relations from low dimension codings.}
\label{fig:model_arch}
\end{figure}

% challenges a system to recover the 
% missing part of an incomplete triple, as a test of its ability to reason from known facts. 
% For example, the system may find statistical evidence suggesting that entities joined by 
% the relation \texttt{adjoining\_country} are also likely to be 
% a tail entity of the relation \texttt{country\_of\_film}, and from the known triple 
% $\langle$\textit{Australia}, \texttt{adjoining\_country}, \textit{New Zealand}$\rangle$, 
% it may imply that \textit{New Zealand} is a probable candidate to fill 
% in the incomplete triple. 

As a most common approach \citep{DBLP:journals/tkde/WangMWG17}, modeling entities and relations to operate in a low dimension 
vector space helps KBC, for three conceivable reasons. First, when dimension is low, entities modeled 
as vectors are forced to share parameters, so ``similar'' entities which participate in 
many relations in common get close to each other (e.g. \textit{Australia} close to \textit{US}). This 
could imply that an entity (e.g. \textit{US}) ``type matches'' a relation such 
as \texttt{country\_of\_film}. Second, relations may share parameters as well, which could transfer 
facts from one relation to other similar relations, for example 
from $\langle$\textit{x}, \texttt{award\_winner}, \textit{y}$\rangle$ to 
$\langle$\textit{x}, \texttt{award\_nominated}, \textit{y}$\rangle$. Third, spatial positions might be 
used to implement \emph{composition} of relations, as relations can be regarded as mappings from head to 
tail entities, and the composition of two maps can match a third (e.g. the composition 
of \texttt{currency\_of\_country} and \texttt{country\_of\_film} 
matches the relation \texttt{currency\_of\_film\_budget}), which 
could be captured by modeling composition in a space. 

However, modeling relations as mappings naturally requires more parameters -- a general linear 
map between $d$-dimension vectors is represented by a matrix of $d^2$ parameters -- 
which are less likely to be shared, impeding transfers of facts between similar relations. 
Thus, it is desired to reduce dimensionality of relations; furthermore, 
the existence of a composition of two relations 
(assumed to be modeled by matrices $\mat{M}_1,\mat{M}_2$) matching a third ($\mat{M}_3$) also justifies 
dimension reduction, because it implies 
a \emph{compositional constraint} $\mat{M}_1\cdot \mat{M}_2\approx \mat{M}_3$ that can be 
satisfied only by a 
lower dimension sub-manifold in the parameter 
space\footnote{It is noteworthy that similar compositional constraints apply to 
most modeling schemes of relations, not just matrices.}. 

Previous approaches reduce dimensionality of relations by imposing 
pre-designed hard constraints on the parameter space, such as constraining that 
relations are translations \citep{DBLP:conf/nips/BordesUGWY13} or diagonal matrices \citep{Yang2015}, or 
assuming they are linear combinations of a small number of prototypes \citep{xie-EtAl:2017:Long}.
However, pre-designed hard constraints do not seem to cope well with compositional constraints, because 
it is difficult to know \emph{a priori} 
which two relations compose to which third relation, hence difficult to choose a pre-design; and 
compositional constraints are not always exact (e.g. the composition 
of \texttt{currency\_of\_country} and \texttt{headquarter\_location} usually matches 
\texttt{business\_operation\_currency} but not always), so hard constraints are 
less suited.

In this paper, we investigate an alternative approach by 
training relation parameters jointly with an autoencoder (Figure~\ref{fig:model_arch}). 
During training, the autoencoder tries to 
reconstruct relations from low dimension codings, with the 
reconstruction objective back-propagating to relation parameters 
as well. We show this novel technique promotes 
parameter sharing between different relations, and drives them toward 
low dimension manifolds (Sec.\ref{sec:analyzeautoenc}). Besides, 
we expect the technique to cope better with compositional constraints, because 
it discovers low dimension manifolds 
posteriorly from data, and it does not impose any explicit hard constraints.

Yet, joint training with an autoencoder is not simple; one has to keep a subtle balance between 
gradients of the reconstruction and KB-learning objectives throughout the training process. 
We are not aware of any theoretical principles directly addressing this problem; but we found some 
important settings after extensive pre-experiments (Sec.\ref{sec:optimizationtricks}). We evaluate our system using standard 
KBC datasets, achieving state-of-the-art on several of them (Sec.\ref{sec:mainresults}), with 
strongly improved Mean Rank. 
We discuss 
detailed settings that lead to the performance (Sec.\ref{sec:trainingbase}), 
and we show that joint training with 
an autoencoder indeed helps discovering 
compositional constraints (Sec.\ref{sec:compositionalconstraints}) and benefits from 
compositional training (Sec.\ref{sec:gainscomptrain}).

\section{Base Model}\label{sec:basemodel}

%In this section we describe a basic linear model for learning KBs. 
A knowledge base (KB) is a set $\setoftriples$ of triples of the 
form $\langle h, r, t\rangle$, where 
$h, t\in\setofents$ are entities and $r\in\setofrels$ is a relation 
(e.g. $\langle$\textit{The Matrix}, \texttt{country\_of\_film}, 
\textit{Australia}$\rangle$). A relation $r$ has its inverse 
$r^{-1}\in\setofrels$ so that for every 
$\langle h, r, t\rangle\in\setoftriples$, 
we regard $\langle t, r^{-1}, h \rangle$ as also in the KB. 
Under this assumption and given $\setoftriples$ as training data, 
we consider the Knowledge Base Completion (KBC) task that 
predicts candidates for a missing tail entity in an incomplete 
$\langle h, r, ?\rangle$ triple. 

Most approaches tackle this problem by training a \emph{score function} 
measuring the plausibility of triples being facts. 
The model we implement in this work represents entities 
$h,t$ as $d$-dimension vectors $\vec{u}_h,\vec{v}_t$ respectively, 
and relation $r$ as a $d\times d$ matrix $\mat{M}_r$. If 
$\vec{u}_h,\vec{v}_t$ are one-hot vectors with 
dimension $d=\lvert\setofents\rvert$ corresponding to each entity, 
one can take $\mat{M}_r$ as the adjacency matrix of entities joined by relation $r$, 
so the set of tail entities filling into $\langle h, r, ?\rangle$ is 
calculated by 
$\vec{u}_h^\top \mat{M}_{r}$ (with each nonzero entry corresponds to an answer). 
Thus, we have $\vec{u}_h^\top \mat{M}_{r}\vec{v}_t > 0$ if and only 
if $\langle h, r, t\rangle\in\setoftriples$. This motivates us to use 
$\vec{u}_h^\top \mat{M}_{r}\vec{v}_t$ as a natural parameter to model plausibility 
of $\langle h, r, t\rangle$, even in a 
%and we expect it to be learned using 
low dimension space with $d\ll\lvert\setofents\rvert$. 
Thus, we define the score function as 
\begin{equation}\label{eq:scrbilinear}
s(h,r,t):=\exp(\vec{u}_h^\top\mat{M}_{r}\vec{v}_t)
\end{equation}
for the basic model. This is similar to the bilinear model 
of \citet{Nickel:2011:TMC:3104482.3104584}, except that we distinguish 
$\vec{u}_h$ (the vector for head entities) from $\vec{v}_t$ (the vector for tail 
entities). It has also been proposed in \citet{tian-okazaki-inui:2016:P16-1}, but 
for modeling dependency trees rather than KBs. 

More generally, we consider \emph{composition} of 
relations $r_1/\ldots/r_l$ to model \emph{path}s in a 
KB \citep{guu-miller-liang:2015:EMNLP}, as defined by 
$r_1,\ldots,r_l$ participating in a 
sequence of facts such that the head entity of 
each fact coincides with the tail of its previous. 
For example, a sequence of two facts 
$\langle$\textit{The Matrix}, \texttt{country\_of\_film}, 
\textit{Australia}$\rangle$ and $\langle$\textit{Australia}, \texttt{currency\_of\_country}, \textit{Australian Dollar}$\rangle$ form a path of composition \texttt{country\_of\_film}\,/ \texttt{currency\_of\_country}, 
because the head of the second 
fact (i.e. \textit{Australia}) coincides with the tail of the first. 
Using the previous $d=\lvert\setofents\rvert$ analogue, one can verify 
that composition of 
relations is represented by multiplication of adjacency matrices, 
so we accordingly define 
$$
s(h,r_1/\ldots/r_l,t):=\exp(\vec{u}_h^\top\mat{M}_{r_1}\cdots\mat{M}_{r_l}\vec{v}_t)
$$
to measure the plausibility of a path. It is explored in 
\citet{guu-miller-liang:2015:EMNLP} to learn a score function not only for 
single facts but also for paths. This \emph{compositional training} scheme 
is shown to bring valuable information about the 
structure of the KB and may help KBC. In this work, we conduct experiments both with and 
without compositional training.

% one can consider \emph{path}s \citep{DBLP:conf/emnlp/GuML15} of the 
% form $\langle h,r_1,\ldots,r_l,t\rangle$, besides triples. A score function measures 
% a path as the plausibility of reaching $t$ from $h$, by traversing a sequence of 
% relations $r_1,\ldots,r_l$ through a \emph{composition} of facts 
% (i.e., a sequence of facts with each head entity coincides with the tail of its 
% previous). For example, one 
% can reach \textit{Australian Dollar} from \textit{The Matrix}, via a composition of 
% the two facts $\langle$\textit{The Matrix}, \texttt{country\_of\_film}, 
% \textit{Australia}$\rangle$ and $\langle$\textit{Australia}, \texttt{currency\_of\_country}, 
% \textit{Australian Dollar}$\rangle$. Note that the head of the second 
% fact (i.e. \textit{Australia}) coincides with the tail of the first. It is shown 
% in \citet{DBLP:conf/emnlp/GuML15} that paths can provide valuable information 
% about structures of KBs and help KBC. In this work, we define 

% ... compositional training ...
% as the score function for paths. This model is similar to the 
% ``compositionalization'' of \citet{Nickel:2011:TMC:3104482.3104584} 
% as described in \citet{DBLP:conf/emnlp/GuML15}, except that we distinguish 
% $\vec{u}_h$ (the vector for head entities) from $\vec{v}_t$ (the vector for tail 
% entities). It has also been proposed in \citet{tian-okazaki-inui:2016}, but for 
% modeling dependency trees rather than KBs. 

In order to learn parameters $\vec{u}_h,\vec{v}_t,\mat{M}_r$ of the score 
function, we 
follow \citet{tian-okazaki-inui:2016:P16-1} using a 
Noise Contrastive Estimation (NCE) \citep{DBLP:journals/jmlr/GutmannH12} objective. For each 
path (or triple) $\langle h,r_1/\ldots,t\rangle$ taken from the KB, 
we generate negative samples by replacing the tail entity $t$ with some 
random noise 
$t^{*}$. Then, we maximize 
\begin{multline*}
\mathcal{L}_1:=
\sum_{\text{path}}\ln\frac{s(h, r_1/\ldots, t)}{k+s(h, r_1/\ldots, t)}\\
+\sum_{\text{noise}}\ln\frac{k}{k+s(h, r_1/\ldots, t^{*})}
\end{multline*}
as our \emph{KB-learning objective}. Here, $k$ is the number of noises 
generated for each path. When the score function 
is regarded as probability, $\mathcal{L}_1$ represents 
the log-likelihood of 
``$\langle h,r_1/\ldots,t\rangle$ being actual path and 
$\langle h,r_1/\ldots,t^{*}\rangle$ being noise''. Maximizing $\mathcal{L}_1$ 
increases the scores of actual paths and decreases the scores of noises.

\section{Joint Training with an Autoencoder}
\label{sec:jointtraining}

% \subsection{Knowledge Base Embedding Model}

% Our KB embedding model is based on the \emph{bilinear comp} \citep{DBLP:conf/emnlp/GuML15}
% and the training scheme of vector-based DCS \citep{tian-okazaki-inui:2016}.

% 知識ベース補完のモデルとして，我々はGuuらのパス情報を取り入れた
% 双線型モデル\cite{DBLP:conf/emnlp/GuML15}とvecDCS\cite{tian-okazaki-inui:2016}の訓練手法をベースにする．この
% モデルでは，エンティティを$d$-次元のベクトル，関係を$(d\times d)$の行列として表現する．
% 学習は，一つのエンティティ$h$から出発し，いくつかの
% 関係$r_1,\ldots,r_n$からなるパスを経由して辿り着いたもう一つの
% エンティティ$t$に対して，エネルギー関数$f(h, r_1,\ldots,r_n, t)$を最大化することで
% エンティティベクトル$\vec{h},\;\vec{t}$と
% 行列$\mat{M}_{r_1},\ldots,\mat{M}_{r_n}$を推定する．
% エネルギー関数は
% \begin{equation}
% f(h, r_1,\ldots,r_n, t) := \exp(^\top\vec{h}\mat{M}_{r_1},\ldots,\mat{M}_{r_n}\vec{t})
% \end{equation}
% と定義し，知識ベースからこのようなデータが取れる尤度に相当する．
% 推定時は，知識ベースからのデータと合わせ，ランダムに生成された$k$個の負例$h, r'_1,\ldots,r'_n, t'$を使って
% \begin{equation}\label{eq:main}
% \frac{f(h, r_1,\ldots,r_n, t)}{k+f(h, r_1,\ldots,r_n, t)}
% \cdot\prod_{k}\frac{k}{k+f(h, r'_1,\ldots,r'_n, t')}
% \end{equation}
% を最大化する．これは，「$(h, r_1,\ldots,r_n, t)$が正例で$(h, r'_1,\ldots,r'_n, t')$が負例」であるイベントの尤度に相当する．

% \subsection{Joint training with an autoencoder}

Autoencoders learn efficient codings of high-dimensional data while trying to 
reconstruct the original data from the coding. By joint training relation matrices 
with an autoencoder, we also expect it to 
help reducing the dimensionality of the original data (i.e. relation matrices). 

Formally, we define a \emph{vectorization} $\vec{m}_r$ for each relation matrix 
$\mat{M}_r$, and use it as input to the autoencoder. $\vec{m}_r$ is defined as a reshape 
of $\mat{M}_r$ flattened into a $d^2$-dimension vector, and normalized such 
that $\lVert\vec{m}_r\rVert=\sqrt{d}$. We define 
\begin{equation}\label{eq:coding}
\vec{c}_r:=\relu(\mat{A}\vec{m}_{r})
\end{equation}
as the coding. Here $\mat{A}$ is a $c\times d^2$ matrix with $c\ll d^2$, and 
$\relu$ is the Rectified Linear Unit function \citep{DBLP:conf/icml/NairH10}. 
We reconstruct the input from $\vec{c}_r$ by multiplying a 
$d^2\times c$ matrix $\mat{B}$. We want $\mat{B}\vec{c}_{r}$ to be more similar 
to $\vec{m}_r$ than other relations. For this purpose, we define a similarity  
\begin{equation}
\label{eq:reconscore}
g(r_1, r_2):=\exp(\frac{1}{\sqrt{dc}}\vec{m}_{r_1}^\top\mat{B}\vec{c}_{r_2}), 
\end{equation}
which measures the length of $\mat{B}\vec{c}_{r_2}$ projected to the direction 
of $\vec{m}_{r_1}$. In order to 
learn the parameters $\mat{A},\mat{B}$, we adopt the Noise Contrastive Estimation scheme as in Sec.\ref{sec:basemodel}, 
generate random noises $r^{*}$ for each relation $r$ and maximize 
$$
\mathcal{L}_2:=
\sum_{r\in\setofrels}\ln\frac{g(r, r)}{k+g(r, r)}
+\sum_{r^{*}\sim\setofrels}\ln\frac{k}{k+g(r, r^{*})}
$$
as our \emph{reconstruction objective}. Maximizing $\mathcal{L}_2$ increases 
$\vec{m}_r$'s similarity with $\mat{B}\vec{c}_{r}$, and decreases it with 
$\mat{B}\vec{c}_{r^{*}}$. 

%The factor $\frac{1}{\sqrt{dc}}$ in $\eqref{eq:reconscore}$ is crucial for 
%decent training, as we will discuss in Sec....

During joint training, both $\mathcal{L}_1$ and $\mathcal{L}_2$ are 
simultaneously maximized, and the gradient $\nabla\mathcal{L}_2$ propagates to 
relation matrices as well. Since $\nabla\mathcal{L}_2$ depends on $\mat{A}$ 
and $\mat{B}$, and 
$\mat{A},\mat{B}$ interact with all relations, they promote indirect 
parameter sharing between different relation matrices. 
In Sec.\ref{sec:analyzeautoenc}, we further 
show that joint training drives relations toward a low dimension manifold.

% Our autoencoder is designed to reconstruct projection matrices of relations.
% Let $\vec{m}_r$ be a $d^2$-dimensional vector
% flattened a projection matrix of relations $\mat{M}_r$.
% The autoencoder transforms $\vec{m}_r$ into a $l$-dimensional code vector ($l \ll d^2$)
% by a matrix $\mat{A} \in \mathbb{R}^{d^2 \times l}$,
% then applies a nonlinear function $\relu$ to the code vector $\mat{A}\vec{m}_r$,
% and reconstructs the input vector by a matrix $\mat{B} \in \mathbb{R}^{l \times d^2}$:
% \begin{equation}
% \vec{m}_{r}\approx\mat{B}\relu(\mat{A}\vec{m}_{r})
% \end{equation}
% We define a scoring function of the autoencoder to maximize the similarity
% between $\vec{m}_r$ and the reconstructed vector as follows.
% \begin{equation}
% g(\vec{m}_{r}):=\exp(\vec{m}_{r}\cdot\mat{B}\relu(\mat{A}\vec{m}_{r}))
% \end{equation}
% を定義し，最適化の際にランダムに生成された$k$個の負例$\vec{m}_{r'}$と合わせて
% \begin{equation}\label{eq:autoencoder}
% \frac{g(\vec{m}_{r})}{k+g(\vec{m}_{r})}\cdot\prod_{k}\frac{k}{k+g(\vec{m}_{r'})}
% \end{equation}
% を最大化する．また，式\eqref{eq:autoencoder}の最大化において$\vec{m}_{r}$に対する
% 勾配も計算し，これと式\eqref{eq:main}で計算された$\vec{m}_{r}$の勾配と合わせて
% パラメータ$\vec{m}_{r}$の更新を行う．オートエンコーダ
% との同時学習によって，$\vec{m}_{r}$が低次元のコードから「復元されやすい位置」，つまり
% 類似した関係同士がクラスタしているような空間位置に動くと期待される。また，全ての関係$r$に
% 対して$\vec{m}_{r}$が同じ行列$\mathbf{A},\;\mathbf{B}$によってエンコード・デコード
% されるので，異なる$\vec{m}_{r}$同士が行列$\mathbf{A},\;\mathbf{B}$を介してパラメータを共有しているとの見方もできる．これによって，異なる関係間の知識共有が促されると思われる．

\section{Optimization Tricks}\label{sec:optimizationtricks}

Joint training with an autoencoder is not simple. Relation matrices receive updates 
from both $\nabla\mathcal{L}_1$ and $\nabla\mathcal{L}_2$, but if they update 
$\nabla\mathcal{L}_1$ too much, the autoencoder has no effect; conversely, if they 
update $\nabla\mathcal{L}_2$ too often, all relation matrices crush into one 
cluster. Furthermore, an autoencoder should learn from 
genuine patterns of relation matrices that emerge from fitting the KB, but not 
the reverse -- in which the autoencoder imposes arbitrary patterns to 
relation matrices 
according to random initialization. Therefore, it is not surprising that 
a naive optimization of 
$\mathcal{L}_1+\mathcal{L}_2$ does not work. 

After extensive pre-experiments, we have found some crucial settings for successful 
training. The most important ``magic'' is the scaling factor $\frac{1}{\sqrt{dc}}$ 
in definition of the similarity function \eqref{eq:reconscore}, perhaps being 
combined with other settings as we discuss below. We have tried different 
factors $1$, $\frac{1}{\sqrt{d}}$, $\frac{1}{\sqrt{c}}$ and 
$\frac{1}{dc}$ instead, with various combinations of $d$ and $c$; but 
the autoencoder failed to learn meaningful codings in other settings. 
When the scaling factor is too small 
(e.g. $\frac{1}{dc}$), all relations get almost the same coding; conversely if 
the factor is too large (e.g. $1$), all codings get very close to $0$. 

The next important rule is to keep a balance between the updates coming from 
$\nabla\mathcal{L}_1$ and $\nabla\mathcal{L}_2$. We 
use Stochastic Gradient 
Descent (SGD) for optimization, and the common practice \citep{bottou2012stochastic} is to set the 
learning rate as
\begin{equation}\label{eq:commonpractice}
\alpha(\tau):=\frac{\eta}{1+\eta\lambda\tau}. 
\end{equation}
Here, $\eta,\lambda$ are hyper-parameters and $\tau$ is a counter of 
processed data points. In this work, in order to control 
the updates in detail to keep a balance, we modify \eqref{eq:commonpractice} to use a 
a step 
counter $\tau_r$ for each relation $r$, counting ``number of updates'' 
instead of 
data points\footnote{Similarly, we set separate step counters for all head 
and tail entities, and the autoencoder as well.}. That is, whenever $\mat{M}_r$ gets a 
nonzero update from a gradient calculation, $\tau_r$ increases by $1$. 
Furthermore, we use different hyper-parameters for different ``types of updates'', 
namely $\eta_1,\lambda_1$ for updates coming from $\nabla\mathcal{L}_1$, and 
$\eta_2,\lambda_2$ for updates coming from $\nabla\mathcal{L}_2$. 
Thus, let $\Delta_1$ be the partial gradient of $\nabla\mathcal{L}_1$, and $\Delta_2$ the partial gradient of $\nabla\mathcal{L}_2$, we 
update $\mat{M}_r$ by $\alpha_1(\tau_r)\Delta_1+\alpha_2(\tau_r)\Delta_2$ at 
each step, where 
$$
\alpha_1(\tau_r):=\frac{\eta_1}{1+\eta_1\lambda_1\tau_r},\;\;
\alpha_2(\tau_r):=\frac{\eta_2}{1+\eta_2\lambda_2\tau_r}.
$$

The rule for setting $\eta_1,\lambda_1$ and $\eta_2,\lambda_2$ 
is that, $\eta_2$ should be much smaller than $\eta_1$, because 
$\eta_1,\eta_2$ control the magnitude of learning rates at the early stage of 
training, with the autoencoder still largely random and $\Delta_2$ not 
making much sense; on the other hand, one has to choose $\lambda_1$ and 
$\lambda_2$ such that 
$\lVert\Delta_1\rVert/\lambda_1$ and $\lVert\Delta_2\rVert/\lambda_2$ are at 
the same scale, because the learning rates approach $1/(\lambda_1\tau_r)$ 
and $1/(\lambda_2\tau_r)$ respectively, as the training proceeds. 
In this way, the autoencoder will not impose random patterns to relation matrices 
according to its initialization at the early stage, and a balance 
is kept between $\alpha_1(\tau_r)\Delta_1$ and $\alpha_2(\tau_r)\Delta_2$ later.

But how to estimate $\lVert\Delta_1\rVert$ and $\lVert\Delta_2\rVert$? It seems 
that we can approximately calculate their scales from initialization. In this 
work, we use i.i.d. Gaussians of variance $1/d$ to initialize parameters, 
so the initial Euclidean norms are $\lVert\vec{u}_h\rVert\approx 1$, 
$\lVert\vec{v}_t\rVert\approx 1$, 
$\lVert\mat{M}_r\rVert\approx\sqrt{d}$, and 
$\lVert\mat{B}\mat{A}\vec{m}_r\rVert\approx\sqrt{dc}$.
%\footnote{We use the Frobenius norm for matrices.}
Thus, by calculating $\nabla\mathcal{L}_1$ and $\nabla\mathcal{L}_2$ using 
\eqref{eq:scrbilinear} and \eqref{eq:reconscore}, we have approximately 
\begin{gather}
\lVert\Delta_1\rVert\approx\lVert\vec{u}_h\vec{v}_t^\top\rVert\approx 1, 
\quad\text{and}\\
\lVert\Delta_2\rVert\approx\lVert\frac{1}{\sqrt{dc}}\mat{B}\vec{c}_r\rVert\approx
\frac{1}{\sqrt{dc}}\lVert\mat{B}\mat{A}\vec{m}_r\rVert\approx 1.
\end{gather}
It suggests that, because of the scaling factor $\frac{1}{\sqrt{dc}}$ in 
\eqref{eq:reconscore}, we have 
$\lVert\Delta_1\rVert$ and $\lVert\Delta_2\rVert$ at the same scale, so we can 
set $\lambda_1=\lambda_2$. This might not be a mere coincidence. 

% In our experiments, the factor $\frac{1}{\sqrt{dc}}$ turns out to be crucial; we 
% have tried using factors $1$, $\frac{1}{\sqrt{d}}$, $\frac{1}{\sqrt{c}}$ and 
% $\frac{1}{dc}$ instead, 
% but although we can achieve balanced joint training each time by adjusting 
% $\lambda_1$ and $\lambda_2$, the autoencoder failed to 
% learn meaningful codings after all. When the factor is too small 
% (e.g. $\frac{1}{dc}$), all relations get almost the same coding; conversely if 
% the factor is too large (e.g. $1$), all codings get very close to $0$. 

% maybe for discussion ???
% This modification of SGD has 
% traits similar to some modern optimization algorithms such as Adagrad \citep{..}, 
% in that they both set different learning rates for different parameters. While 
% Adagrad sets them adaptively by keeping track of past gradients for 
% all parameters, our modification of SGD is more efficient and allows us to 
% grasp a rough intuition about which parameter gets how much update. 

\subsection{Training the Base Model}\label{sec:trainingbase}

Besides the tricks for joint training, we also found settings that 
significantly improve the base model on KBC, as briefly 
discussed below. In Sec.\ref{sec:crucialsettings}, we will show 
performance gains by these settings using the FB15k-237 validation set.

\paragraph{Normalization} It is better to normalize 
relation matrices to $\lVert\mat{M}_r\rVert=\sqrt{d}$ during training. 
This might reduce 
fluctuations in entity vector updates. 

\paragraph{Regularizer} It is better to minimize 
$\lVert \mat{M}_r^\top \mat{M}_r-\frac{1}{d}\tr(\mat{M}_r^\top \mat{M}_r)I\rVert$ during training. 
This regularizer drives $\mat{M}_r$ toward an orthogonal matrix 
\citep{tian-okazaki-inui:2016:P16-1} and might reduce 
fluctuations in entity vector updates. As a result, all relation matrices trained 
in this work are very close to orthogonal.

\paragraph{Initialization} Instead of pure Gaussian, it is better to initialize 
matrices as $(I+G)/2$, where $G$ is random. The identity matrix $I$ helps passing information from 
head to tail \citep{tian-okazaki-inui:2016:P16-1}.

\paragraph{Negative Sampling} Instead of a unigram distribution, it is better 
to use a \emph{uniform} distribution for generating noises. This is 
somehow counter-intuitive compared to training word embeddings. 

%\subsection{Training of Autoencoder}
\section{Related Works}

KBs have a wide range of applications
\citep{berant-EtAl:2013:EMNLP,hixon-clark-hajishirzi:2015:NAACL-HLT,DBLP:journals/pieee/Nickel0TG16}
and KBC has inspired a huge amount of research 
\citep{DBLP:conf/nips/BordesUGWY13,riedel-EtAl:2013:NAACL-HLT,DBLP:conf/nips/SocherCMN13,DBLP:conf/aaai/WangZFC14,wang-EtAl:2014:EMNLP20145,DBLP:conf/ijcai/xiao16,nguyen-EtAl:2016:N16-1,
toutanova-EtAl:2016:P16-1,das-EtAl:2017:EACLlong1,hayashi-shimbo:2017:Short}. 

Among the previous works, TransE \cite{DBLP:conf/nips/BordesUGWY13} is the 
classic method which represents a relation as a translation of the entity 
vector space, and is partially inspired by \citet{mikolov-yih-zweig:2013:NAACL-HLT}'s vector arithmetic method of 
solving word analogy tasks. Although competitive in KBC, it is speculated that 
this method is well-suited for $1$-to-$1$ relations but might be too 
simple to represent $N$-to-$N$ relations 
accurately\cite{DBLP:journals/tkde/WangMWG17}. Thus, extensions such as 
TransR \cite{DBLP:conf/aaai/LinLSLZ15} and STransE \cite{nguyen-EtAl:2016:N16-1} 
are proposed to map entities into a relation-specific vector space before 
translation. The ITransF model \cite{xie-EtAl:2017:Long} further enhances this 
approach by imposing a hard constraint that the relation-specific maps should be 
linear combinations of a small number of prototypical matrices. Our work 
inherits the same motivation with ITransF in terms of 
promoting parameter-sharing among relations. 

On the other hand, the base model used in this work originates from 
RESCAL \cite{Nickel:2011:TMC:3104482.3104584}, in which relations are 
naturally represented as analogue to the adjacency matrices 
(Sec.\ref{sec:basemodel}). Further 
developments include HolE \cite{DBLP:conf/aaai/NickelRP16} and 
ConvE \cite{dettmers2018conve} which improve this approach in terms of 
parameter-efficiency, by introducing low dimension factorizations of the 
matrices. We inherit the basic model of RESCAL but draw 
additional training techniques from \citet{tian-okazaki-inui:2016:P16-1}, and 
show that the base model already can achieve near 
state-of-the-art performance (Sec.\ref{sec:mainresults},\ref{sec:crucialsettings}). This sends a message 
similar to \citet{kadlec-bajgar-kleindienst:2017:RepL4NLP}, saying that 
training tricks might be as important as model designs.

Nevertheless, we emphasize the novelty of this work in that the previous 
models mostly achieve dimension reduction by imposing some pre-designed 
hard constraints \citep{DBLP:conf/nips/BordesUGWY13,Yang2015,DBLP:conf/icml/TrouillonWRGB16,DBLP:conf/aaai/NickelRP16,xie-EtAl:2017:Long,dettmers2018conve}, whereas the 
constraints themselves are not learned from data; in contrast, 
our approach by jointly training an autoencoder does not impose 
any explicit hard constraints, so it leads to more flexible modeling. 

Moreover, we additionally focus on leveraging composition in KBC. Although 
this idea has been frequently explored 
before \citep{guu-miller-liang:2015:EMNLP,neelakantan-roth-mccallum:2015:ACL-IJCNLP,lin-EtAl:2015:EMNLP1}, our discussion about the concept of 
compositional constraints and its connection to dimension reduction 
has not been addressed similarly in previous research. In experiments, we will 
show (Sec.\ref{sec:compositionalconstraints},\ref{sec:gainscomptrain}) 
that joint training 
with an autoencoder indeed helps finding compositional constraints and 
benefits from compositional training.

%In this work, we achieved strong performance with a very simple base 
%model similar to \citet{Nickel:2011:TMC:3104482.3104584}, by applying 
%some detailed but crucial settings. This sends a message similar to 
%\citet{kadlec-bajgar-kleindienst:2017:RepL4NLP}.

%Nevertheless, we have focused on composition, an idea that has been 
%explored frequently 
%\citep{guu-miller-liang:2015:EMNLP,neelakantan-roth-mccallum:2015:ACL-IJCNLP,lin-EtAl:2015:EMNLP1}, but our discussion about the concept of 
%compositional constraints and its connection to dimension reduction 
%has not been addressed similarly in previous research. 

%Dimension reduction is commonly addressed by KB embedding models, mostly with 
%a pre-designed low dimension representation \citep{DBLP:conf/nips/BordesUGWY13,Yang2015,DBLP:conf/icml/TrouillonWRGB16}, 
%or factorization into 
%low dimension spaces \citep{DBLP:conf/aaai/NickelRP16,dettmers2018conve}. The paradigm of learning such dimension reduction 
%from data has been pursued by \citet{xie-EtAl:2017:Long}, but they still impose hard constraints 
%on the parameter space. In contrast, joint training with an autoencoder does not impose 
%any explicit constraints, which could lead to more flexible modeling. 

Autoencoders have been used solo for learning distributed representations 
of syntactic trees \citep{socher-EtAl:2011:EMNLP}, words and 
images \citep{silberer-lapata:2014:P14-1}, 
or semantic roles \citep{titov-khoddam:2015:NAACL-HLT}. It is also used 
for pretraining other deep neural networks \citep{Erhan:2010:WUP}. 
However, when combined with other models, the learning of autoencoders, or more generally 
\emph{sparse codings} \citep{rubinstein2010dictionaries}, is 
usually conveyed in an alternating manner, %by alternatively , 
fixing one part of the model while optimizing the other, 
such as in \citet{xie-EtAl:2017:Long}. 
To our knowledge, joint training with an autoencoder is not widely used previously 
for reducing dimensionality. 

Jointly training an autoencoder is not simple because it takes 
non-stationary inputs. 
In this work, we modified SGD so that it shares 
%have rearranged some common practice of SGD, which has 
traits with some modern optimization algorithms such as Adagrad \citep{DBLP:journals/jmlr/DuchiHS11}, 
in that they both set different learning rates for different parameters. While 
Adagrad sets them adaptively by keeping track of gradients for 
all parameters, our modification of SGD is more efficient and allows us to 
grasp a rough intuition about which parameter gets how much update. 
We believe our techniques and findings in joint training with an autoencoder 
could be helpful to reducing dimensionality and improving interpretability in 
other neural network architectures as well. 

\section{Experiments}

We evaluate on standard KBC datasets, 
%For evaluating our proposed model,
%we use a selection of standard KBC datasets from the literature:
including WN18 and FB15k \citep{DBLP:conf/nips/BordesUGWY13}, 
WN18RR \citep{dettmers2018conve} and FB15k-237 \citep{toutanova-chen:2015:CVSC}.
%WN18 \citep{DBLP:conf/nips/BordesUGWY13},
%FB15k \citep{DBLP:conf/nips/BordesUGWY13},
%WN18RR \citep{dettmers2018conve}, and
%FB15k-237 \citep{Toutanova2015a}.
The statistical information of these datasets are shown 
in Table~\ref{tab:datasets}.

\begin{table}[t]
\centering
\setlength{\tabcolsep}{4pt}
\small
\begin{tabular}{lrrrrr}
\toprule
Dataset & \multicolumn{1}{c}{$\lvert\setofents\rvert$} & \multicolumn{1}{c}{$\lvert\setofrels\rvert$} & \multicolumn{1}{c}{\#Train} & \multicolumn{1}{c}{\#Valid} & \multicolumn{1}{c}{\#Test} \\
\midrule
WN18 & 40,943 & 18 & 141,442 & 5,000 & 5,000 \\
FB15k & 14,951 & 1,345 & 483,142 & 50,000 & 59,071 \\
WN18RR & 40,943 & 11 & 86,835 & 3,034 & 3,134 \\
FB15k-237 & 14,541 & 237 & 272,115 & 17,535 & 20,466 \\
%YAGO3-10 & 123,182 & 37 & 1,079,040 & 5,000 & 5,000 \\
%Countries S1 & 271 & 2 & 1,111 & 24 & 24 \\
%Countries S2 & 271 & 2 & 1,063 & 24 & 24 \\
%Countries S3 & 271 & 2 & 985 & 24 & 24 \\
%Nations & 70 & 14 & 1,619 & 202 & 203 \\
%UMLS & 135 & 46 & 5216 & 652 & 661 \\
%Kinship & 129 & 104 & 8,548 & 1,069 & 1,069 \\
\bottomrule
\end{tabular}
\caption{%
Statistical information of the KBC datasets.
$\lvert\setofents\rvert$ and $\lvert\setofrels\rvert$ denote the number of 
entities and relation types, respectively; 
\#Train, \#Valid, and \#Test are the numbers of triples
in the training, validation, and test sets, respectively.}
\label{tab:datasets}
\end{table}

WN18 collects word relations from 
WordNet \citep{DBLP:journals/cacm/Miller95}, and FB15k is taken from 
Freebase \citep{DBLP:conf/sigmod/BollackerEPST08}; both have filtered out low 
frequency entities. However, it is reported in \citet{toutanova-chen:2015:CVSC} that 
both WN18 and FB15k have information leaks because 
the inverses of some test triples appear in the training set. 
FB15k-237 and WN18RR fix this problem by deleting such triples from training 
and test data. In this work, we do evaluate on WN18 and FB15k, but our models 
are mainly tuned on FB15k-237. 

For all datasets, we set the dimension $d=256$ and $c=16$, the SGD 
hyper-parameters $\eta_1=1/64$, 
$\eta_2=2^{-14}$ and $\lambda_1=\lambda_2=2^{-14}$. The training batch size 
is 32 and the triples in each batch share the same head entity. 
We compare the base model (\textsc{base}) to our joint training 
with an autoencoder model (\textsc{joint}), and the base 
model with compositional training (\textsc{base+comp}) to our
joint model with compositional training (\textsc{joint+comp}). 
When compositional training is 
enabled (\textsc{base+comp}, \textsc{joint+comp}), 
we use random walk to sample paths of length $1+X$, where $X$ is drawn from a Poisson 
distribution of mean $\lambda=1.0$.

% We note that \citet{Toutanova2015a} have reported that
% WN18 and FB15k have an information leak between the training and test sets.
% Although we evaluate our model on these datasets to compare with more previous methods,
% we conduct a detailed analysis of our method on FB15k-237 dataset that do not have such leak.

For any incomplete triple $\mtriple{h}{r}{?}$ in KBC test, we calculate 
a score $s(h,r,e)$ from \eqref{eq:scrbilinear}, for 
every entity $e\in\setofents$ such that $\mtriple{h}{r}{e}$ 
\emph{does not appear in any of the training, validation, or test sets}
\citep{DBLP:conf/nips/BordesUGWY13}. 
Then, the calculated scores together with $s(h,r,t)$ for the gold triple 
is converted to ranks, and the rank of the gold entity $t$ is used for 
evaluation. Evaluation metrics include Mean Rank (MR),
Mean Reciprocal Rank (MRR), and
Hits at 10 (H10). 
Lower MR, higher MRR, and higher H10 indicate better performance.

We consult MR and MRR on validation sets to determine training epochs; we stop 
training when both MR and MRR have stopped improving.

%evaluation metrics
% The knowledge base completion is a task
% to predict the head or the tail entity given the relation and the other entity,
% i.e., predict $h$ given $\mtriple{?}{r}{t}$ or predict $t$ given $\mtriple{h}{r}{?}$.
% Specifically, we use \emph{filtered} evaluation protocol proposed by \citet{DBLP:conf/nips/BordesUGWY13}.
% For all test triples $\mtriple{h}{r}{t}$,
% (1) we calculate scores $s(e, r, t)$ for all triples $\mtriple{e}{r}{t}$
% s.t. $e \in \setofents$ and $\mtriple{e}{r}{t} \not\in \setoftriples \setminus \{\mtriple{h}{r}{t}\}$,\footnote{%
% This is to avoid penalizing the model for ranking other correct triples higher than the testing triple.}
% (2) we sort values by decreasing order, and
% (3) we record the rank of correct triple $\mtriple{h}{r}{t}$.
% A same process is repeated for predicting $t$.
% We report
% the mean of those predicted ranks (MR),
% the mean of reciprocal ranks (MRR), and
% the proportion of correct entities ranked in the top $k$ (Hits@$k$).
% Lower MR, higher MRR, or higher Hits@$k$ mean better performance.

\subsection{KBC Results}\label{sec:mainresults}

\begin{figure}[!t]
\centering
\includegraphics[width=\columnwidth]{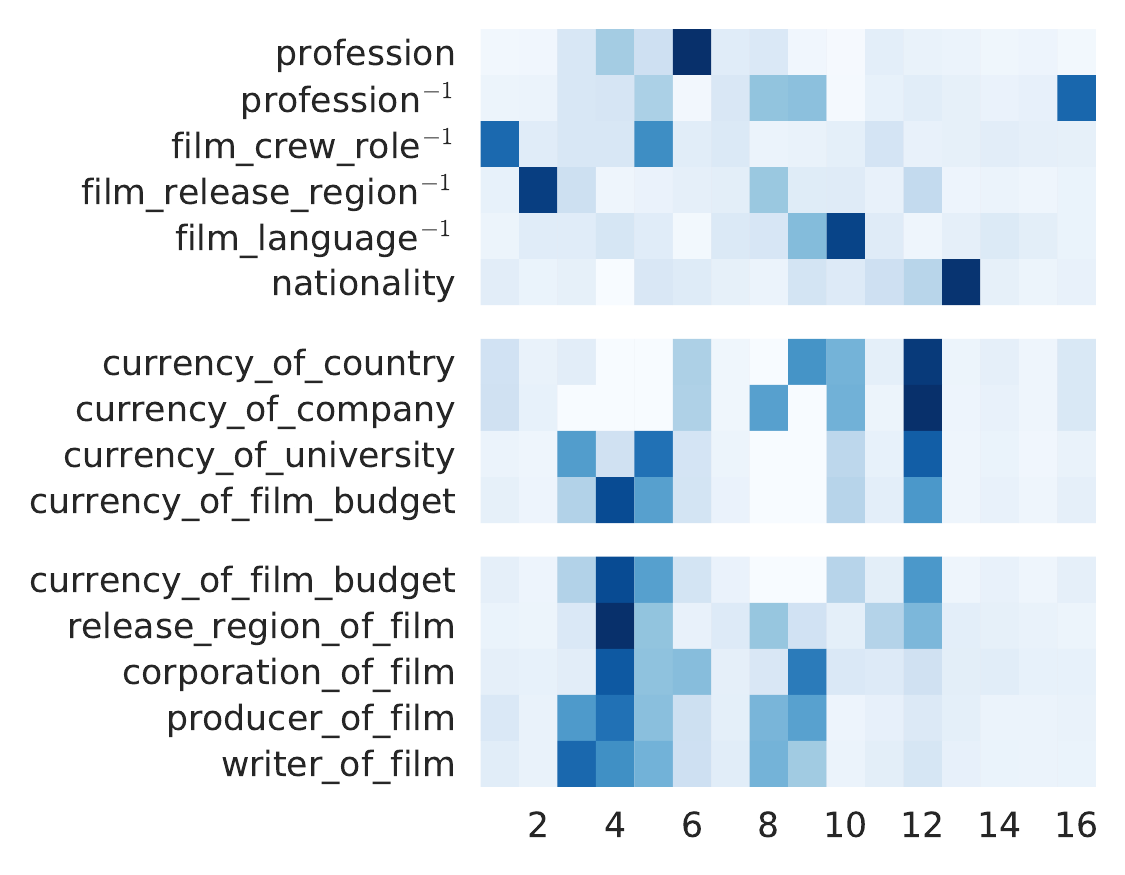}
\caption{%
Examples of relation codings learned from FB15k-237. Each row shows 
a 16 dimension vector encoding a relation. Vectors are normalized 
such that their entries sum to $1$.}
\label{fig:code-heatmap}
\end{figure}

\begin{table*}[!t]
\centering
\setlength{\tabcolsep}{5pt}
\small
\begin{tabular}{@{}lcccccccccc@{}}
\toprule
\multirow{2}{*}{Model} & \multicolumn{2}{c}{WN18} & \multicolumn{2}{c}{FB15k} & \multicolumn{3}{c}{WN18RR} & \multicolumn{3}{c}{FB15k-237} \\
\cmidrule(lr){2-3} \cmidrule(lr){4-5} \cmidrule(lr){6-8} \cmidrule(l){9-11}
 & MR & H10 & MR & H10 & MR & MRR & H10 & MR & MRR & H10 \\
\midrule
%\textsc{joint} & 279 & 95.8 & 53 & 81.9 & \makecell{3514 \\ (3253)} & \makecell{\textbf{.464} \\ (\textbf{.484})} & \makecell{\textbf{55.8} \\ (\textbf{56.7})} & 212 & .336 & \textbf{52.3} \\
%\textsc{base} & 288 & 95.7 & 52 & 82.0 & \makecell{3500 \\ (3193)} & \makecell{.462 \\ (.484)} & \makecell{55.7 \\ (56.6)} & 215 & \textbf{.337} & \textbf{52.3} \\
\textsc{joint} & \textbf{277} & \textbf{95.8} & \textbf{53} & \textbf{82.5} & \textbf{4233} & \textbf{.461}$^*$ & \textbf{53.4} & \textbf{212} & .336 & \textbf{52.3}$^*$ \\
\textsc{base} & 286 & \textbf{95.8} & \textbf{53} & \textbf{82.5} & 4371 & .459 & 52.9 & 215 & \textbf{.337}$^*$ & \textbf{52.3}$^*$ \\
\midrule
%\textsc{joint+comp} & \textbf{191} & 94.8 & 61 & 74.2 & \makecell{2293 \\ (1906)} & \makecell{.286 \\ (.305)} & \makecell{52.2 \\ (55.8)} & \textbf{197} & .331 & 51.6 \\
%\textsc{base+comp} & 195 & 94.8 & 65 & 74.3 & \makecell{\textbf{2214} \\ (\textbf{1853})} & \makecell{.284 \\ (.304)} & \makecell{52.0 \\ (55.6)} & 203 & .328 & 51.5 \\
\textsc{joint+comp} & \textbf{191}$^*$ & \textbf{94.8} & \textbf{53} & \textbf{69.7} & \textbf{2268}$^*$ & \textbf{.343} & \textbf{54.8}$^*$ & \textbf{197}$^*$ & \textbf{.331} & \textbf{51.6} \\
\textsc{base+comp} & 195 & \textbf{94.8} & 54 & 69.4 & 2447 & .310 & 54.1 & 203 & .328 & 51.5 \\
\midrule
%TransE & 433 & 94.3 & 63 & 64.0 & 5749 & .17 & 40.8 & 234 & .27 & 44.6 \\
TransE \citep{DBLP:conf/nips/BordesUGWY13} & 292 & 92.0 & \textbf{66} & 70.4 & 4311 & .202 & 45.6 & \textbf{278} & .236 & 41.6 \\
%TransR \citep{DBLP:conf/aaai/LinLSLZ15} & \textbf{281} & 93.6 & 77 & 68.7 & \textbf{4222} & .210 & \textbf{47.1} & 320 & \textbf{.282} & \textbf{45.9} \\ % ACL2018 submission
TransR \citep{DBLP:conf/aaai/LinLSLZ15} & \textbf{281} & 93.6 & 76 & \textbf{74.4} & \textbf{4222} & .210 & \textbf{47.1} & 320 & \textbf{.282} & \textbf{45.9} \\
RESCAL \citep{Nickel:2011:TMC:3104482.3104584} & 911 & 58.0 & 163 & 41.0 & 9689 & .105 & 20.3 & 457 & .178 & 31.9 \\
HolE \citep{DBLP:conf/aaai/NickelRP16} & 724 & \textbf{94.3} & 293 & 66.8 & 8096 & \textbf{.376} & 40.0 & 1172 & .169 & 30.9 \\
\midrule
%TransE & 251 & 89.2 & 125 & 47.1 & - & - & - & - & - & - \\
%TransH & 303 & - & 86.7 & 87 & - & 64.4 & - & - & - & &&& - & - & - \\
%TransR & 225 & -     & 92.0   & -      & -      & 77  & -     & 68.7   & -      & -      \\
STransE \citep{nguyen-EtAl:2016:N16-1} & 206 & 93.4 & 69 & 79.9 & - & - & - & - & - & - \\
ITransF \citep{xie-EtAl:2017:Long} & \textbf{205} & 94.2 & 65 & 81.0 & - & - & - & - & - & - \\
%HolE & - & 94.9 & - & 73.9 & - & - & - & - & - & - \\
%RESCAL$^\dagger$ & - & 92.8 & - & 58.7 & - & - & - & - & - & - \\
%DistMult$^\ddagger$ & 902 & 93.6 & 97 & 82.4 & 5110 & .43 & 49 & 254 & .241 & 41.9 \\
ComplEx \citep{DBLP:conf/icml/TrouillonWRGB16} & - & 94.7 & - & 84.0 & \textbf{5261} & .44 & \textbf{51} & 339 & .247 & 42.8 \\
%\texttt{fastText} & - & 94.9 & - & 86.5 & - & - & - & - & - & 44.8 \\
%Single DistMult & 655 & 94.6 & 42.2 & 89.3 & - & - & - & - & - & - \\
Ensemble DistMult \citep{kadlec-bajgar-kleindienst:2017:RepL4NLP} & 457 & 95.0 & 35.9 & 90.4 & - & - & - & - & - & - \\
IRN \citep{shen-EtAl:2017:RepL4NLP1} & 249 & 95.3 & 38 & \textbf{92.7$^*$} & - & - & - & - & - & - \\
ConvE \citep{dettmers2018conve} & 504 & 95.5 & 64 & 87.3 & 5277 & \textbf{.46} & 48 & \textbf{246} & \textbf{.316} & \textbf{49.1} \\
%Inverse Model & 567 & 96.9 & 1897 & 73.7 & - & - & - & - & - & - \\
R-GCN+ \citep{DBLP:journals/corr/SchlichtkrullKB17} & - & \textbf{96.4}$^*$ & - & 84.2 & - & - & - & - & .249 & 41.7 \\
%\texttt{fastText} - train+valid & - & \textbf{97.6} & - & 89.9 & - & - & - & - & - & 45.8 \\
ProjE \citep{DBLP:conf/aaai/ShiW17} & - & - & \textbf{34$^*$} & 88.4 & - & - & - & - & - & - \\
\bottomrule
\end{tabular}
\caption{%
% Knowledge base completion results on WN18, FB15k, WN18RR, and FB15k-237.
% H10 stands for Hits@10.
% Results mark (${}^\dagger$) and (${}^\ddagger$) taken from
% \citet{DBLP:conf/aaai/NickelRP16} and \citet{dettmers2018conve}, respectively.
% The number in the brackets denote the result of ignoring test triples that contain OOV entities.
% The first and second block represents our models with and without compositional training, respectively.
% The third block includes previous models.
KBC results on the WN18, FB15k, WN18RR, and FB15k-237 datasets. The first and 
second sectors compare our joint to the base models with and without compositional 
training, respectively; the third sector shows our re-experiments and the fourth 
shows previous published results. Bold numbers are the best in each sector, and $(^*)$ indicates 
the best of all.}
\label{tab:main-results}
\end{table*}

The results are shown in Table~\ref{tab:main-results}. We found that joint 
training with an autoencoder mostly improves performance, and 
the improvement becomes more clear when compositional training 
is enabled (i.e., $\textsc{joint}\geq\textsc{base}$ and $\textsc{joint+comp}>\textsc{base+comp}$). This is convincing because 
generally, joint training contributes with its regularizing effects, 
and drastic improvements are less 
expected\footnote{The source code and trained models are 
publicly released at \url{https://github.com/tianran/glimvec}, 
where we also show the mean performance and deviations of 
multiple random initializations, to give a more 
complete picture.}.
When compositional training is enabled, the system usually 
achieves better MR, though not always improves in other 
measures. The performance gains are more obvious on the 
WN18RR and FB15k-237 datasets, possibly because WN18 and 
FB15k contain a lot of easy instances that can be solved 
by a simple rule \cite{dettmers2018conve}.

%to performance by its 
%regularizing effects, 
%the joint training technique works more or less like a regularizer, 
%so we do not expect drastic gains. % in performance. 

Furthermore, the numbers demonstrated by our joint and base models are among 
the strongest in the literature. We have conducted re-experiments of several representative 
algorithms, and also compare with state-of-the-art published results. 
For re-experiments, we use \citet{DBLP:conf/aaai/LinLSLZ15}'s 
implementation\footnote{\url{https://github.com/thunlp/KB2E}} of 
TransE \citep{DBLP:conf/nips/BordesUGWY13} and TransR, 
which represent relations as vector translations; and \citet{DBLP:conf/aaai/NickelRP16}'s 
implementation\footnote{\url{https://github.com/mnick/holographic-embeddings}} of 
RESCAL \citep{Nickel:2011:TMC:3104482.3104584} and HolE, where RESCAL is most similar to 
the \textsc{base} model and HolE is a more parameter-efficient variant. We experimented with the default settings, 
and found that our models outperform most of them. 

Among the published results, STransE \citep{nguyen-EtAl:2016:N16-1} and ITransF \citep{xie-EtAl:2017:Long} are more complicated versions of TransR, achieving the previous highest MR on WN18 
but are outperformed by our \textsc{joint+comp} model. ITransF is most similar to 
our \textsc{joint} model in that they both learn sparse codings for relations. 
On WN18RR and FB15k-237, \citet{dettmers2018conve}'s report of 
ComplEx \citep{DBLP:conf/icml/TrouillonWRGB16} and ConvE were previously the best results. 
Our models mostly outperform them. 
Other results 
include \citet{kadlec-bajgar-kleindienst:2017:RepL4NLP}'s simple but strong baseline and several recent 
models 
\citep{DBLP:journals/corr/SchlichtkrullKB17,DBLP:conf/aaai/ShiW17,shen-EtAl:2017:RepL4NLP1} which achieve best results on FB15k or WN18 in some measure. Our models have comparable results.

\subsection{Intuition and Insight}\label{sec:analyzeautoenc}

What does the autoencoder look like? How does joint training affect 
relation matrices? We address these questions by analyses 
showing that \textbf{(i)} the autoencoder learns sparse and interpretable 
codings of relations, \textbf{(ii)} the joint training drives relation matrices 
toward a low dimension manifold, and \textbf{(iii)} it helps discovering compositional 
constraints.

\subsubsection*{Sparse Coding and Interpretability}\label{sec:interpretability}

Due to the $\relu$ function in \eqref{eq:coding}, our autoencoder learns 
sparse coding, with 
most relations having large code values at only two or three dimensions. 
This sparsity makes it easy to find patterns in the 
model that to some extent explain the semantics of relations.
Figure~\ref{fig:code-heatmap} shows 
some examples. 

In the first group of Figure~\ref{fig:code-heatmap}, we show a small number 
of relations that are almost always assigned a near one-hot coding, 
regardless of initialization. 
%Firstly, regardless of initialization, ... a fixed set of relations that ...
%These are shown as the first group of Figure...
These are high frequency relations joining two large categories 
(e.g. film and language), which probably constitute the skeleton of a KB. 

In the second group, we found the $12$th dimension strongly correlates with 
\texttt{currency}; and in the third group, we found the $4$th dimension 
strongly correlates with \texttt{film}. As for the relation 
\texttt{currency\_of\_film\_budget}, it has large code values at both dimensions. 
This kind of relation clustering also seems independent of initialization. 
Intuitively, it shows that the autoencoder may discover similarities 
between relations and 
promote indirect parameter sharing among them. 
Yet, as the autoencoder only reconstructs \emph{approximations} of relation 
matrices but never constrain them to be exactly equal to the original, relation 
matrices with very similar codings may still differ considerably. For 
example, 
\texttt{producer\_of\_film} and \texttt{writer\_of\_film} have
codings of cosine similarity 
0.973, but their relation matrices only 
have\footnote{Cosine similarity 0.338 is still high for matrices, due to the high 
dimensionality of their parameter space.} a 
cosine similarity 0.338.

\subsubsection*{Low dimension manifold}

\begin{figure}[!t]
\centering
\begin{subfigure}[b]{0.48\columnwidth}
\centering
\includegraphics[width=\textwidth]{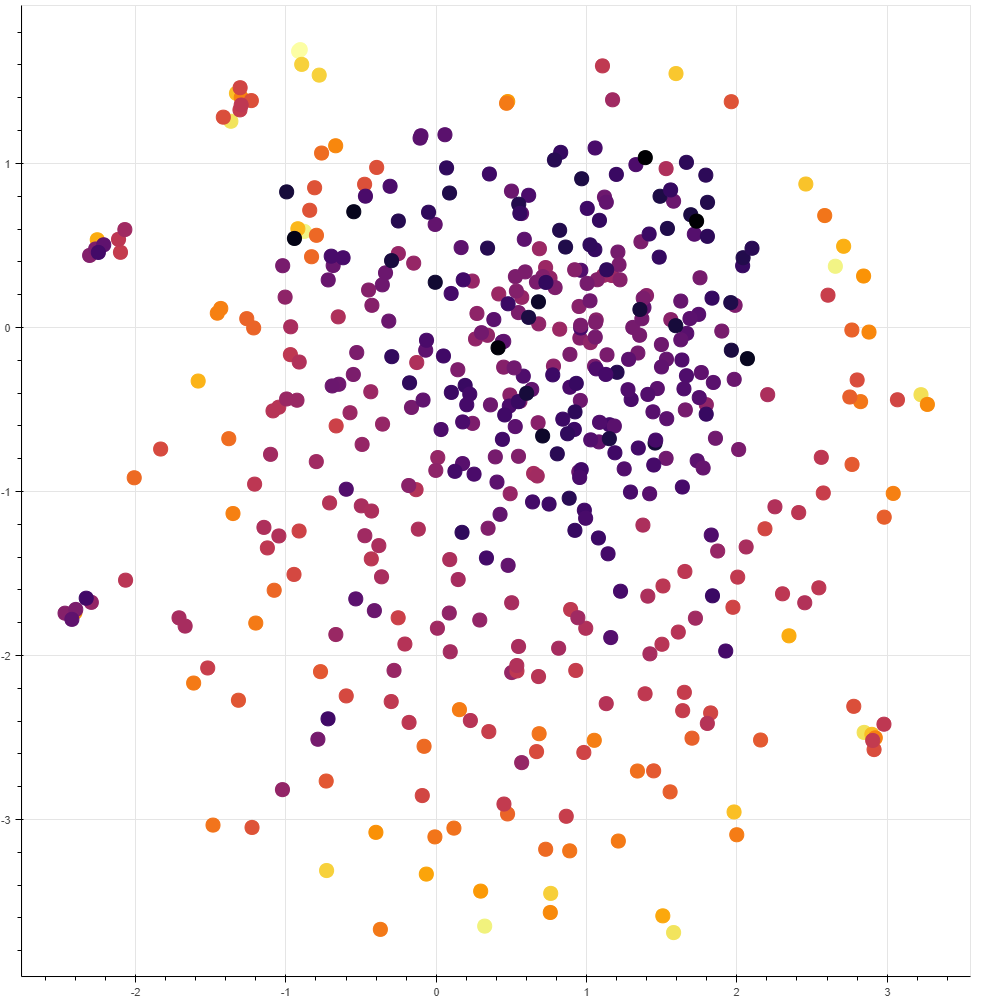}
% http://www.cl.ecei.tohoku.ac.jp/~ryo-t/dcsveckb/model-nobr-noautoenc-nocomp-2048_umap.html
\caption{\textsc{base}}
\label{subfig:umap-base}
\end{subfigure}
\begin{subfigure}[b]{0.48\columnwidth}
\centering
\includegraphics[width=\textwidth]{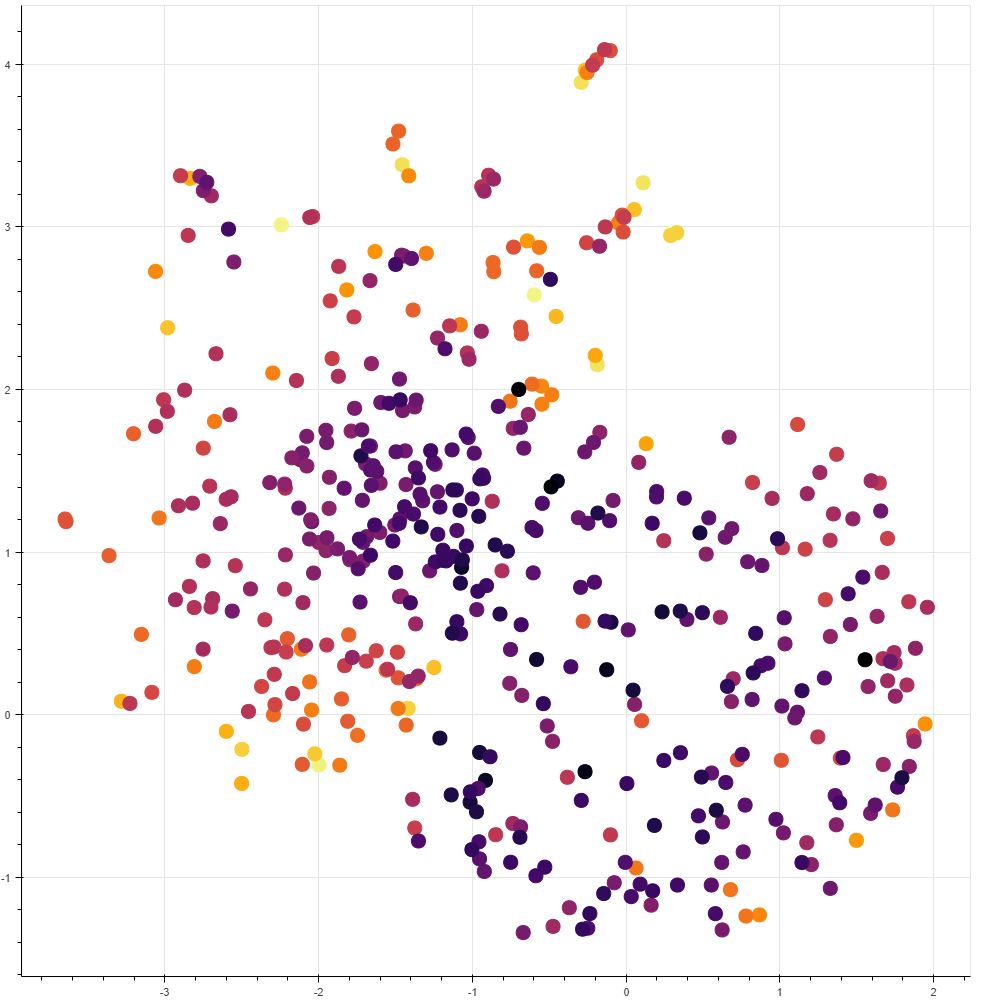}
% http://www.cl.ecei.tohoku.ac.jp/~ryo-t/dcsveckb/model-nobr-nocomp-2048_umap.html
\caption{\textsc{joint}}
\label{subfig:umap-joint}
\end{subfigure}
\begin{subfigure}[b]{0.48\columnwidth}
\centering
\includegraphics[width=\textwidth]{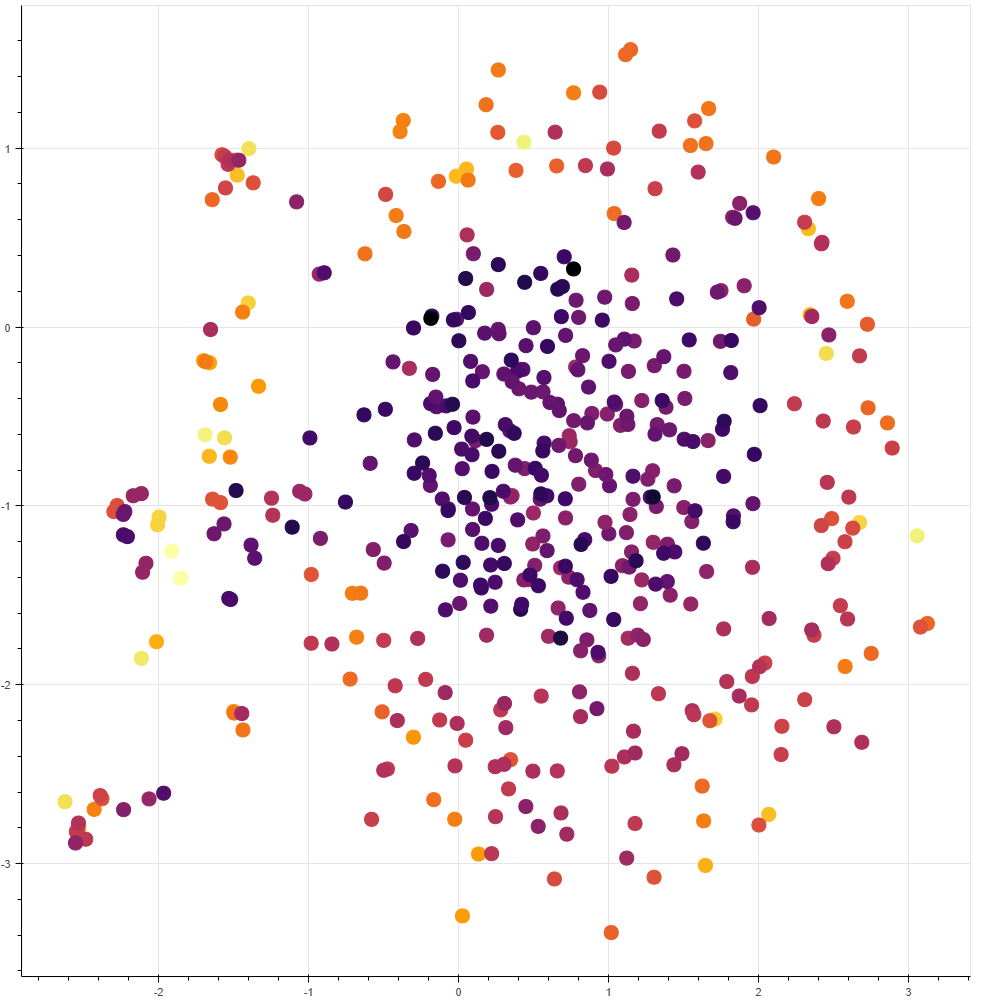}
% http://www.cl.ecei.tohoku.ac.jp/~ryo-t/dcsveckb/model-nobr-noautoenc-nolex-1.0-2048_umap.html
\caption{\textsc{base+comp}}
\label{subfig:umap-base+comp}
\end{subfigure}
\begin{subfigure}[b]{0.48\columnwidth}
\centering
\includegraphics[width=\textwidth]{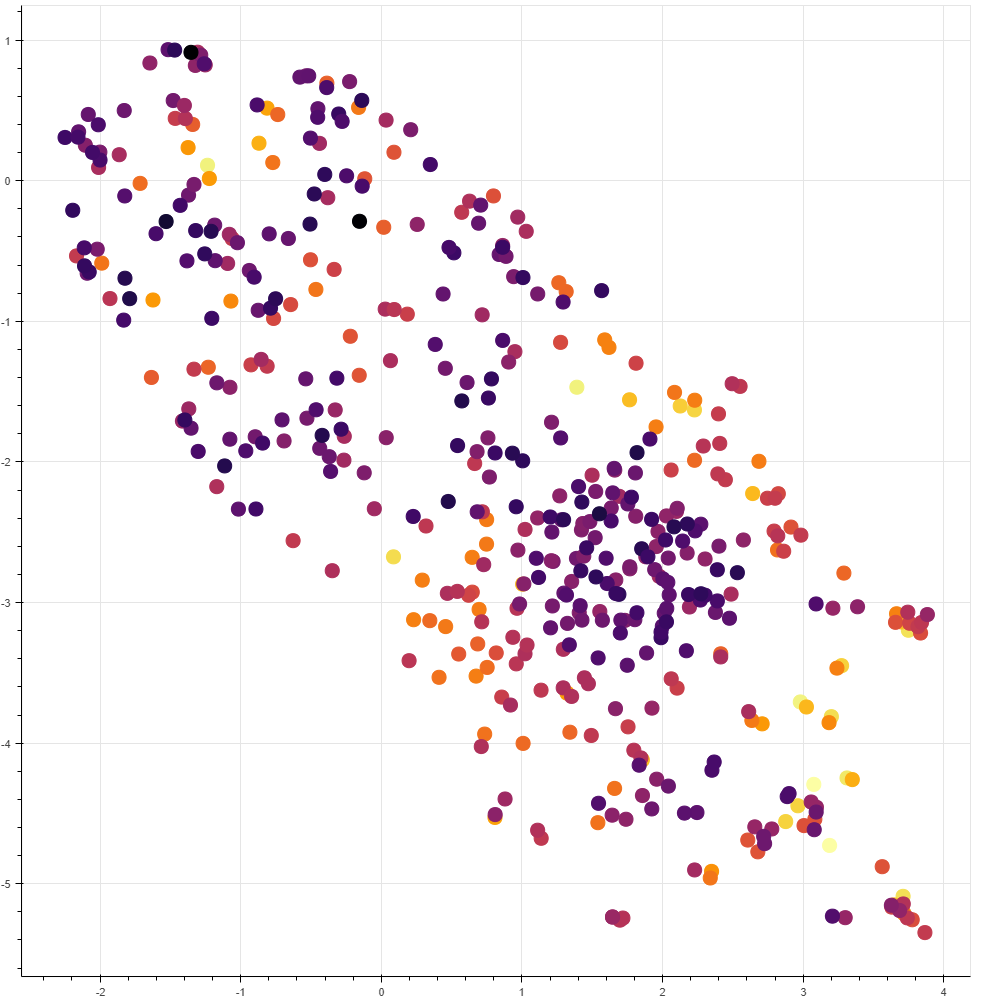}
% http://www.cl.ecei.tohoku.ac.jp/~ryo-t/dcsveckb/model-nobr-nolex-1.0-2048_umap.html
\caption{\textsc{joint+comp}}
\label{subfig:umap-joint+comp}
\end{subfigure}
\caption{%
By UMAP, relation matrices are embedded into a 2D plane. 
%Visualization of relation matrices by UMAP. Each dot is a high dimension matrix 
%projected to the 2D plane. 
Colors show frequencies of relations; and lighter color means more frequent.}
%Comparison of UMAP embeddings for relations matrices between \textsc{base+comp} and the \textsc{joint+comp}.
%The darker the color, the lower the frequency of the relation in the training set.}
\label{fig:umap}
\end{figure}

In order to visualize the relation matrices learned by our joint and base 
models, we use 
UMAP\footnote{\url{https://github.com/lmcinnes/umap}} 
\citep{2018arXivUMAP} to 
embed $\mat{M}_r$ into a 2D 
plane\footnote{UMAP is a recently proposed manifold 
learning algorithm based on the fuzzy topological structure. 
We also tried 
t-SNE \citep{maaten2008visualizing} but found UMAP more insightful.}. 
We use relation matrices trained on FB15k-237, and compare models trained by 
the same number of epochs. 
The results are shown in Figure~\ref{fig:umap}. 

We can see that Figure~\ref{subfig:umap-base} and Figure~\ref{subfig:umap-base+comp} are mostly similar, with high frequency 
relations scattered randomly around a low frequency cluster, suggesting that 
they come from various directions of a high dimension space, with 
frequent relations probably being pulled further by the training updates. 
On the other hand, in Figure~\ref{subfig:umap-joint} and Figure~\ref{subfig:umap-joint+comp} we found less frequent relations being 
clustered with frequent ones, and multiple traces of low dimension 
structures. It suggests that joint training with an autoencoder 
indeed drives relations toward a low dimension manifold. In addition, 
Figure~\ref{subfig:umap-joint+comp} shows different structures against Figure~\ref{subfig:umap-joint}, which we conjecture could be 
related to compositional constraints discovered by compositional training.

% To show that joint training drives relations toward a low-dimensional manifold,
% in Figure~\ref{fig:umap}
% we plot relation matrices in the 2D space using UMAP \citep{2018arXivUMAP},
% which is a non-linear dimension reduction technique similar to t-SNE \citep{maaten2008visualizing}.

% In \textsc{base+comp} (Figure~\ref{subfig:noautoenc}),
% lower frequency relations form a large cluster in the center of Figure~\ref{subfig:noautoenc}
% and higher frequency relations are spread out radially around the cluster.
% In \textsc{joint+comp} (Figure~\ref{subfig:full-model}),
% lower frequency relations still form a large cluster,
% but some parts of the cluster spreads away from the center,
% and whole relations are arranged into 潰れた感じに．
% This suggests that ...

% Note that these observations were valid across different initialization values and UMAP parameters (APPENDIX).

\subsubsection*{Compositional constraints}\label{sec:compositionalconstraints}

In order to directly evaluate a model's ability to find compositional constraints, 
we extracted from FB15k-237 a list of $(r_1/r_2, r_3)$ pairs such that 
$r_1/r_2$ matches $r_3$. Formally, the list is constructed as below. 
%built a dataset from FB15k-237 as follows. 
%if joint training with an autoencoder helps discovering compositional constraints, 
For any relation $r$, we define 
a \emph{content set} $C(r)$ as the set of $(h,t)$ pairs such that 
$\langle h,r,t\rangle$ is a fact in the KB. Similarly, we define 
$C(r_1/r_2)$ as the set of $(h,t)$ pairs such that 
$\langle h,r_1/r_2,t\rangle$ is a path.
We regard $(r_1/r_2, r_3)$ as a 
compositional constraint if their content sets are similar; 
that is, 
if $\lvert C(r_1/r_2)\cap C(r_3) \rvert\geq 50$ and 
the Jaccard similarity between $C(r_1/r_2)$ and $C(r_3)$ is $\geq 0.4$. Then, after 
filtering out degenerated cases such as $r_1=r_3$ or $r_2=r_1^{-1}$, we 
obtained a list of 154 compositional constraints, e.g. \\
(\texttt{currency\_of\_country}/\texttt{country\_of\_film}, 
\texttt{currency\_of\_film\_budget}). 

\begin{table}[!t]
\centering
\small
\setlength{\tabcolsep}{14pt}
\begin{tabular}{lrr}
\toprule
Model & MR & MRR \\
\midrule
\textsc{joint+comp} & \textbf{130$\pm$27} & \textbf{.0481$\pm$.0090} \\ 
\textsc{base+comp} & 150$\pm$3 & .0280$\pm$.0010 \\ 
\textsc{RandomM2} & 181$\pm$19 & .0356$\pm$.0100 \\
\bottomrule
\end{tabular}
\caption{Performance at discovering compositional constraints 
extracted from FB15k-237}
\label{tab:compositional-constraints}
\end{table}

For each compositional constraint $(r_1/r_2, r_3)$ in the list, 
we take the matrices $\mat{M}_1$, $\mat{M}_2$ and $\mat{M}_3$ 
corresponding to $r_1$, $r_2$ and $r_3$ respectively, and 
rank $\mat{M}_3$ according to its cosine similarity with $\mat{M}_1\mat{M}_2$, among all relation 
matrices. Then, we calculate MR and MRR for evaluation. 
We compare the \textsc{joint+comp} model 
to \textsc{base+comp}, as well as a randomized baseline where 
$M_2$ is selected randomly from the relation matrices in \textsc{joint+comp} instead (\textsc{RandomM2}).
The results are shown in Table~\ref{tab:compositional-constraints}. 
We have evaluated 5 different random initializations for each model, trained 
by the same 
number of epochs, and we report the mean and standard deviation. We verify 
that \textsc{joint+comp} performs better than \textsc{base+comp}, 
indicating that 
joint training with an autoencoder indeed helps discovering 
compositional constraints. Furthermore, the random baseline 
\textsc{RandomM2} tests a hypothesis that joint training might be just clustering $M_3$ and $M_1$ here, to the extent that 
$M_3$ and $M_1$ are so close that even 
a random $M_2$ can give the correct answer; but as it turns out, \textsc{joint+comp} largely 
outperforms \textsc{RandomM2}, excluding this possibility. Thus, 
joint training performs better not simply because 
it clusters relation matrices; it learns compositions indeed.

\subsection{Losses and Gains}

In the KBC task, where are the losses and what are the gains of 
different settings?
With additional evaluations, we show \textbf{(i)} some crucial 
settings for the base model, and 
\textbf{(ii)} joint training with an autoencoder benefits more from
compositional training. 
%joint training gains more as it focuses more on composition. %, 
%and \textbf{(iii)} the losses of compositional training. 

\subsubsection*{Crucial settings for the base model}\label{sec:crucialsettings}

\begin{table}[!t]
\centering
\small
\setlength{\tabcolsep}{8pt}
\begin{tabular}{rrrr}
\toprule
\multicolumn{1}{c}{Settings} & MR & MRR & H10 \\
\midrule
\multicolumn{1}{l}{\textsc{base}} & \textbf{214} & \textbf{.338} & \textbf{52.5} \\ % model-nobr-noautoenc-nocomp-512
\midrule
no normalization & 309 & .326 & 49.9 \\ % model-nonorm-512
no regularizer & 400 & .328 & 51.3 \\ % model-weakreg-512
pure Gaussian & 221 & .336 & 52.1 \\ % model-gaussinit-512
\qquad unigram distribution& 215 & .324 & 50.6 \\ % model-unigramnegsamp-512
\bottomrule
\end{tabular}
\caption{Ablation of the four settings of the base model as described 
in Sec.\ref{sec:trainingbase}}
\label{tab:crucial-settings}
\end{table}

It is noteworthy that our base model already achieves strong results. 
This is due to several detailed but crucial settings 
as we discussed in Sec.\ref{sec:trainingbase}; 
Table~\ref{tab:crucial-settings}
shows their gains on 
the FB15k-237 validation data. 
The most dramatic improvement comes from the regularizer that drives matrices 
to orthogonal. 

% To reveal crucial settings leading to our state-of-the-art results,
% We examine how performance varies given different initialization on FB15k-237 (Table~\ref{tab:crucial-settings}).
% \begin{table}[!t]
% \centering
% \setlength{\tabcolsep}{3pt}
% \begin{tabular}{@{}lccccc@{}}
% \toprule
% \multirow{2}{*}{Method} & & & \multicolumn{3}{c}{Hits} \\
% \cmidrule(l){4-6}
% & MR & MRR & @10 & @3 & @1 \\
% \midrule
% Proposed & 194 & .334 & 51.9 & 36.7 & 24.2 \\ % model-nobr-nolex-1.0-512
% init2 & 192 & .335 & 51.8 & 36.7 & 24.4 \\ % init2/model-nobr-nolex-1.0-512
% init3 & 197 & .334 & 51.9 & 36.8 & 24.2 \\ % init3/model-nobr-nolex-1.0-512
% init4 & 195 & .335 & 52.1 & 36.7 & 24.3 \\ % init4/model-nobr-nolex-1.0-512
% gaussinit & 221 & .336 & 52.1 & 36.7 & 24.5 \\ % model-gaussinit-512
% unigramnegsamp & 215 & .324 & 50.6 & 35.2 & 23.4 \\ % model-unigramnegsamp-512
% weakreg & 400 & .328 & 51.3 & 36.3 & 23.5 \\ % model-weakreg-512
% nonorm & 309 & .326 & 49.9 & 35.8 & 23.8 \\  % model-nonorm-512
% \bottomrule
% \end{tabular}
% \caption{Validation set, FB15k-237, same epochs}
% \label{tab:crucial-settings}
% \end{table}

\subsubsection*{Gains with compositional training}\label{sec:gainscomptrain}

One can force a model to focus more on (longer) compositions of relations, 
by sampling longer paths in compositional training. 
Since joint training with an autoencoder helps discovering compositional 
constraints, we expect it to be more helpful when the sampled paths are longer. 
In this work, path lengths are sampled from a Poisson distribution, we thus 
vary the mean $\lambda$ of the Poisson to control the strength of compositional 
training. The results on FB15k-237 are shown in Table~\ref{tab:ablation-ae-comp}. 

We can see that, as $\lambda$ gets larger, MR improves much but MRR slightly drops. 
It suggests that in FB15k-237, composition of relations might mainly help 
finding more appropriate candidates for a missing entity, rather than 
pinpointing a correct one. Yet, joint training improves base models even more 
as the paths get longer, especially in MR. 
It further supports our conjecture that joint training with an autoencoder 
may strongly interact with compositional training. 

%\section{Losses of compositional training}

%However, in our experiments ...

\begin{table}[!t]
\centering
\setlength{\tabcolsep}{5pt}
\small
\begin{tabular}{@{}lcrrrrrr@{}}
\toprule
\multirow{2}{*}{Model} & \multirow{2}{*}{$\lambda$} & \multicolumn{3}{c}{Valid} & \multicolumn{3}{c}{Test} \\
\cmidrule(lr){3-5} \cmidrule(l){6-8}
& & MR & MRR & H10 & MR & MRR & H10 \\
\midrule
\textsc{base}  & 0 & 209 & .341 & 52.9 & 215 & .337 & 52.3 \\ % model-nobr-noautoenc-nocomp-3584
\textsc{joint} & 0 & +1 & -.001 & -.2 & \textbf{-3} & -.001 & 0 \\ % model-nobr-nocomp-2048
\midrule
\textsc{base}  & 0.5 & 204 & .337 & 52.2 & 211 & .332 & 51.7 \\ % model-nobr-noautoenc-nolex-0.5-1024
\textsc{joint} & 0.5 & \textbf{-3} & \textbf{+.002} & \textbf{+.1} & +1 & \textbf{+.002} & \textbf{+.2} \\ % model-nobr-nolex-0.5-512
\midrule
\textsc{base}  & 1.0 & 191 & .334 & 52.0 & 203 & .328 & 51.5 \\ % model-nobr-noautoenc-nolex-1.0-1536
\textsc{joint} & 1.0 & \textbf{-5} & \textbf{+.002} & -.1 & \textbf{-6} & \textbf{+.003} & \textbf{+.1} \\ % model-nobr-nolex-1.0-2560
\bottomrule
\end{tabular}
\caption{%
Evaluation of \textsc{base} and gains by \textsc{joint}, on FB15k-237 
with different strengths of compositional training. Bold numbers are improvements.}
\label{tab:ablation-ae-comp}
\end{table}

\section{Conclusion}

We have investigated a dimension reduction technique which trains a KB embedding model jointly with an autoencoder. 
%In this paper, we have trained embedding models for knowledge bases (KBs) and 
%investigated a technique which reduces the dimensionality of relation parameters 
%by jointly training with an autoencoder. 
We have developed new training techniques and achieved 
state-of-the-art results on 
several KBC tasks with strong improvements in Mean Rank. 
Furthermore, we have shown that the autoencoder learns low dimension sparse 
codings that can be easily explained; the joint training technique 
drives high-dimensional data toward low dimension manifolds; and the 
reduction of dimensionality may interact strongly 
with composition, help discovering compositional constraints 
and benefit from compositional training. 
We believe these findings provide insightful understandings of KB embedding models and might be applied to other neural networks beyond the KBC task.

\section*{Acknowledgments}

This work was supported by JST CREST Grant Number JPMJCR1301, Japan. We thank 
Pontus Stenetorp, Makoto Miwa, and the anonymous reviewers for many helpful 
advices and comments.

\bibliographystyle{acl_natbib}
\bibliography{acl2018}

\begin{thebibliography}{45}
\expandafter\ifx\csname natexlab\endcsname\relax\def\natexlab#1{#1}\fi

\bibitem[{Auer et~al.(2007)Auer, Bizer, Kobilarov, Lehmann, Cyganiak, and
  Ives}]{auer2007dbpedia}
S{\"{o}}ren Auer, Christian Bizer, Georgi Kobilarov, Jens Lehmann, Richard
  Cyganiak, and Zachary~G. Ives. 2007.
\newblock \href {https://doi.org/10.1007/978-3-540-76298-0_52} {Dbpedia: {A}
  nucleus for a web of open data}.
\newblock In \emph{The Semantic Web, 6th International Semantic Web Conference,
  2nd Asian Semantic Web Conference, {ISWC} 2007 + {ASWC} 2007, Busan, Korea,
  November 11-15, 2007.}, pages 722--735.

\bibitem[{Berant et~al.(2013)Berant, Chou, Frostig, and
  Liang}]{berant-EtAl:2013:EMNLP}
Jonathan Berant, Andrew Chou, Roy Frostig, and Percy Liang. 2013.
\newblock \href {http://www.aclweb.org/anthology/D13-1160} {Semantic parsing on
  {Freebase} from question-answer pairs}.
\newblock In \emph{Proceedings of the 2013 Conference on Empirical Methods in
  Natural Language Processing}, pages 1533--1544, Seattle, Washington, USA.
  Association for Computational Linguistics.

\bibitem[{Bollacker et~al.(2008)Bollacker, Evans, Paritosh, Sturge, and
  Taylor}]{DBLP:conf/sigmod/BollackerEPST08}
Kurt~D. Bollacker, Colin Evans, Praveen Paritosh, Tim Sturge, and Jamie Taylor.
  2008.
\newblock \href {https://doi.org/10.1145/1376616.1376746} {Freebase: a
  collaboratively created graph database for structuring human knowledge}.
\newblock In \emph{Proceedings of the {ACM} {SIGMOD} International Conference
  on Management of Data, {SIGMOD} 2008, Vancouver, BC, Canada, June 10-12,
  2008}, pages 1247--1250.

\bibitem[{Bordes et~al.(2013)Bordes, Usunier, Garc{\'{\i}}a{-}Dur{\'{a}}n,
  Weston, and Yakhnenko}]{DBLP:conf/nips/BordesUGWY13}
Antoine Bordes, Nicolas Usunier, Alberto Garc{\'{\i}}a{-}Dur{\'{a}}n, Jason
  Weston, and Oksana Yakhnenko. 2013.
\newblock \href
  {http://papers.nips.cc/paper/5071-translating-embeddings-for-modeling-multi-relational-data}
  {Translating embeddings for modeling multi-relational data}.
\newblock In \emph{Advances in Neural Information Processing Systems 26: 27th
  Annual Conference on Neural Information Processing Systems 2013. Proceedings
  of a meeting held December 5-8, 2013, Lake Tahoe, Nevada, United States.},
  pages 2787--2795.

\bibitem[{Bottou(2012)}]{bottou2012stochastic}
L{\'e}on Bottou. 2012.
\newblock \href {https://doi.org/10.1007/978-3-642-35289-8_25} {Stochastic
  gradient descent tricks}.
\newblock In \emph{Neural Networks: Tricks of the Trade}, pages 421--436.
  Springer.

\bibitem[{Das et~al.(2017)Das, Neelakantan, Belanger, and
  McCallum}]{das-EtAl:2017:EACLlong1}
Rajarshi Das, Arvind Neelakantan, David Belanger, and Andrew McCallum. 2017.
\newblock \href {http://www.aclweb.org/anthology/E17-1013} {Chains of reasoning
  over entities, relations, and text using recurrent neural networks}.
\newblock In \emph{Proceedings of the 15th Conference of the European Chapter
  of the Association for Computational Linguistics: Volume 1, Long Papers},
  pages 132--141, Valencia, Spain. Association for Computational Linguistics.

\bibitem[{Dettmers et~al.(2018)Dettmers, Pasquale, Pontus, and
  Riedel}]{dettmers2018conve}
Tim Dettmers, Minervini Pasquale, Stenetorp Pontus, and Sebastian Riedel. 2018.
\newblock \href {https://arxiv.org/abs/1707.01476} {Convolutional 2d knowledge
  graph embeddings}.
\newblock In \emph{Proceedings of the 32th AAAI Conference on Artificial
  Intelligence}.

\bibitem[{Duchi et~al.(2011)Duchi, Hazan, and
  Singer}]{DBLP:journals/jmlr/DuchiHS11}
John~C. Duchi, Elad Hazan, and Yoram Singer. 2011.
\newblock \href {http://dl.acm.org/citation.cfm?id=2021068} {Adaptive
  subgradient methods for online learning and stochastic optimization}.
\newblock \emph{Journal of Machine Learning Research}, 12:2121--2159.

\bibitem[{Erhan et~al.(2010)Erhan, Bengio, Courville, Manzagol, Vincent, and
  Bengio}]{Erhan:2010:WUP}
Dumitru Erhan, Yoshua Bengio, Aaron~C. Courville, Pierre{-}Antoine Manzagol,
  Pascal Vincent, and Samy Bengio. 2010.
\newblock \href {https://doi.org/10.1145/1756006.1756025} {Why does
  unsupervised pre-training help deep learning?}
\newblock \emph{Journal of Machine Learning Research}, 11:625--660.

\bibitem[{Gutmann and Hyv{\"{a}}rinen(2012)}]{DBLP:journals/jmlr/GutmannH12}
Michael Gutmann and Aapo Hyv{\"{a}}rinen. 2012.
\newblock \href {http://dl.acm.org/citation.cfm?id=2188396} {Noise-contrastive
  estimation of unnormalized statistical models, with applications to natural
  image statistics}.
\newblock \emph{Journal of Machine Learning Research}, 13:307--361.

\bibitem[{Guu et~al.(2015)Guu, Miller, and Liang}]{guu-miller-liang:2015:EMNLP}
Kelvin Guu, John Miller, and Percy Liang. 2015.
\newblock \href {http://aclweb.org/anthology/D15-1038} {Traversing knowledge
  graphs in vector space}.
\newblock In \emph{Proceedings of the 2015 Conference on Empirical Methods in
  Natural Language Processing}, pages 318--327, Lisbon, Portugal. Association
  for Computational Linguistics.

\bibitem[{Hayashi and Shimbo(2017)}]{hayashi-shimbo:2017:Short}
Katsuhiko Hayashi and Masashi Shimbo. 2017.
\newblock \href {http://aclweb.org/anthology/P17-2088} {On the equivalence of
  holographic and complex embeddings for link prediction}.
\newblock In \emph{Proceedings of the 55th Annual Meeting of the Association
  for Computational Linguistics (Volume 2: Short Papers)}, pages 554--559,
  Vancouver, Canada. Association for Computational Linguistics.

\bibitem[{Hixon et~al.(2015)Hixon, Clark, and
  Hajishirzi}]{hixon-clark-hajishirzi:2015:NAACL-HLT}
Ben Hixon, Peter Clark, and Hannaneh Hajishirzi. 2015.
\newblock \href {http://www.aclweb.org/anthology/N15-1086} {Learning knowledge
  graphs for question answering through conversational dialog}.
\newblock In \emph{Proceedings of the 2015 Conference of the North American
  Chapter of the Association for Computational Linguistics: Human Language
  Technologies}, pages 851--861, Denver, Colorado. Association for
  Computational Linguistics.

\bibitem[{Kadlec et~al.(2017)Kadlec, Bajgar, and
  Kleindienst}]{kadlec-bajgar-kleindienst:2017:RepL4NLP}
Rudolf Kadlec, Ondrej Bajgar, and Jan Kleindienst. 2017.
\newblock \href {http://www.aclweb.org/anthology/W17-2609} {Knowledge base
  completion: Baselines strike back}.
\newblock In \emph{Proceedings of the 2nd Workshop on Representation Learning
  for NLP}, pages 69--74, Vancouver, Canada. Association for Computational
  Linguistics.

\bibitem[{Lin et~al.(2015{\natexlab{a}})Lin, Liu, Luan, Sun, Rao, and
  Liu}]{lin-EtAl:2015:EMNLP1}
Yankai Lin, Zhiyuan Liu, Huanbo Luan, Maosong Sun, Siwei Rao, and Song Liu.
  2015{\natexlab{a}}.
\newblock \href {http://aclweb.org/anthology/D15-1082} {Modeling relation paths
  for representation learning of knowledge bases}.
\newblock In \emph{Proceedings of the 2015 Conference on Empirical Methods in
  Natural Language Processing}, pages 705--714, Lisbon, Portugal. Association
  for Computational Linguistics.

\bibitem[{Lin et~al.(2015{\natexlab{b}})Lin, Liu, Sun, Liu, and
  Zhu}]{DBLP:conf/aaai/LinLSLZ15}
Yankai Lin, Zhiyuan Liu, Maosong Sun, Yang Liu, and Xuan Zhu.
  2015{\natexlab{b}}.
\newblock \href
  {https://www.aaai.org/ocs/index.php/AAAI/AAAI15/paper/view/9571} {Learning
  entity and relation embeddings for knowledge graph completion}.
\newblock In \emph{Proceedings of the Twenty-Ninth {AAAI} Conference on
  Artificial Intelligence, January 25-30, 2015, Austin, Texas, {USA.}}, pages
  2181--2187.

\bibitem[{van~der Maaten and Hinton(2008)}]{maaten2008visualizing}
Laurens van~der Maaten and Geoffrey Hinton. 2008.
\newblock \href {http://www.jmlr.org/papers/v9/vandermaaten08a.html}
  {Visualizing data using {t-SNE}}.
\newblock \emph{Journal of Machine Learning Research}, 9:2579--2605.

\bibitem[{{McInnes} and {Healy}(2018)}]{2018arXivUMAP}
L.~{McInnes} and J.~{Healy}. 2018.
\newblock \href {http://arxiv.org/abs/1802.03426} {{UMAP: Uniform Manifold
  Approximation and Projection for Dimension Reduction}}.
\newblock \emph{ArXiv e-prints}.

\bibitem[{Mikolov et~al.(2013)Mikolov, Yih, and
  Zweig}]{mikolov-yih-zweig:2013:NAACL-HLT}
Tomas Mikolov, Wen-tau Yih, and Geoffrey Zweig. 2013.
\newblock \href {http://www.aclweb.org/anthology/N13-1090} {Linguistic
  regularities in continuous space word representations}.
\newblock In \emph{Proceedings of the 2013 Conference of the North American
  Chapter of the Association for Computational Linguistics: Human Language
  Technologies}, pages 746--751, Atlanta, Georgia. Association for
  Computational Linguistics.

\bibitem[{Miller(1995)}]{DBLP:journals/cacm/Miller95}
George~A. Miller. 1995.
\newblock \href {https://doi.org/10.1145/219717.219748} {Wordnet: {A} lexical
  database for english}.
\newblock \emph{Commun. {ACM}}, 38(11):39--41.

\bibitem[{Nair and Hinton(2010)}]{DBLP:conf/icml/NairH10}
Vinod Nair and Geoffrey~E. Hinton. 2010.
\newblock \href {http://www.icml2010.org/papers/432.pdf} {Rectified linear
  units improve restricted boltzmann machines}.
\newblock In \emph{Proceedings of the 27th International Conference on Machine
  Learning (ICML-10), June 21-24, 2010, Haifa, Israel}, pages 807--814.

\bibitem[{Neelakantan et~al.(2015)Neelakantan, Roth, and
  McCallum}]{neelakantan-roth-mccallum:2015:ACL-IJCNLP}
Arvind Neelakantan, Benjamin Roth, and Andrew McCallum. 2015.
\newblock \href {http://www.aclweb.org/anthology/P15-1016} {Compositional
  vector space models for knowledge base completion}.
\newblock In \emph{Proceedings of the 53rd Annual Meeting of the Association
  for Computational Linguistics and the 7th International Joint Conference on
  Natural Language Processing (Volume 1: Long Papers)}, pages 156--166,
  Beijing, China. Association for Computational Linguistics.

\bibitem[{Nguyen et~al.(2016)Nguyen, Sirts, Qu, and
  Johnson}]{nguyen-EtAl:2016:N16-1}
Dat~Quoc Nguyen, Kairit Sirts, Lizhen Qu, and Mark Johnson. 2016.
\newblock \href {http://www.aclweb.org/anthology/N16-1054} {Stranse: a novel
  embedding model of entities and relationships in knowledge bases}.
\newblock In \emph{Proceedings of the 2016 Conference of the North American
  Chapter of the Association for Computational Linguistics: Human Language
  Technologies}, pages 460--466, San Diego, California. Association for
  Computational Linguistics.

\bibitem[{Nickel et~al.(2016{\natexlab{a}})Nickel, Murphy, Tresp, and
  Gabrilovich}]{DBLP:journals/pieee/Nickel0TG16}
Maximilian Nickel, Kevin Murphy, Volker Tresp, and Evgeniy Gabrilovich.
  2016{\natexlab{a}}.
\newblock \href {https://doi.org/10.1109/JPROC.2015.2483592} {A review of
  relational machine learning for knowledge graphs}.
\newblock \emph{Proceedings of the {IEEE}}, 104(1):11--33.

\bibitem[{Nickel et~al.(2016{\natexlab{b}})Nickel, Rosasco, and
  Poggio}]{DBLP:conf/aaai/NickelRP16}
Maximilian Nickel, Lorenzo Rosasco, and Tomaso~A. Poggio. 2016{\natexlab{b}}.
\newblock \href {https://dl.acm.org/citation.cfm?id=3016172} {Holographic
  embeddings of knowledge graphs}.
\newblock In \emph{Proceedings of the Thirtieth {AAAI} Conference on Artificial
  Intelligence, February 12-17, 2016, Phoenix, Arizona, {USA.}}, pages
  1955--1961.

\bibitem[{Nickel et~al.(2011)Nickel, Tresp, and
  Kriegel}]{Nickel:2011:TMC:3104482.3104584}
Maximilian Nickel, Volker Tresp, and Hans-Peter Kriegel. 2011.
\newblock \href {http://dl.acm.org/citation.cfm?id=3104482.3104584} {A
  three-way model for collective learning on multi-relational data}.
\newblock In \emph{Proceedings of the 28th International Conference on
  International Conference on Machine Learning}, ICML'11, pages 809--816, USA.
  Omnipress.

\bibitem[{Riedel et~al.(2013)Riedel, Yao, McCallum, and
  Marlin}]{riedel-EtAl:2013:NAACL-HLT}
Sebastian Riedel, Limin Yao, Andrew McCallum, and Benjamin~M. Marlin. 2013.
\newblock \href {http://www.aclweb.org/anthology/N13-1008} {Relation extraction
  with matrix factorization and universal schemas}.
\newblock In \emph{Proceedings of the 2013 Conference of the North American
  Chapter of the Association for Computational Linguistics: Human Language
  Technologies}, pages 74--84, Atlanta, Georgia. Association for Computational
  Linguistics.

\bibitem[{Rubinstein et~al.(2010)Rubinstein, Bruckstein, and
  Elad}]{rubinstein2010dictionaries}
R.~Rubinstein, A.~M. Bruckstein, and M.~Elad. 2010.
\newblock \href {https://doi.org/10.1109/JPROC.2010.2040551} {Dictionaries for
  sparse representation modeling}.
\newblock \emph{Proceedings of the IEEE}, 98(6):1045--1057.

\bibitem[{Schlichtkrull et~al.(2017)Schlichtkrull, Kipf, Bloem, van~den Berg,
  Titov, and Welling}]{DBLP:journals/corr/SchlichtkrullKB17}
Michael~Sejr Schlichtkrull, Thomas~N. Kipf, Peter Bloem, Rianne van~den Berg,
  Ivan Titov, and Max Welling. 2017.
\newblock \href {http://arxiv.org/abs/1703.06103} {Modeling relational data
  with graph convolutional networks}.
\newblock \emph{CoRR}, abs/1703.06103.

\bibitem[{Shen et~al.(2017)Shen, Huang, Chang, and
  Gao}]{shen-EtAl:2017:RepL4NLP1}
Yelong Shen, Po-Sen Huang, Ming-Wei Chang, and Jianfeng Gao. 2017.
\newblock \href {http://www.aclweb.org/anthology/W17-2608} {Modeling
  large-scale structured relationships with shared memory for knowledge base
  completion}.
\newblock In \emph{Proceedings of the 2nd Workshop on Representation Learning
  for NLP}, pages 57--68, Vancouver, Canada. Association for Computational
  Linguistics.

\bibitem[{Shi and Weninger(2017)}]{DBLP:conf/aaai/ShiW17}
Baoxu Shi and Tim Weninger. 2017.
\newblock \href
  {https://www.aaai.org/ocs/index.php/AAAI/AAAI17/paper/viewPaper/14279}
  {Proje: Embedding projection for knowledge graph completion}.
\newblock In \emph{Proceedings of the Thirty-First {AAAI} Conference on
  Artificial Intelligence, February 4-9, 2017, San Francisco, California,
  {USA.}}, pages 1236--1242.

\bibitem[{Silberer and Lapata(2014)}]{silberer-lapata:2014:P14-1}
Carina Silberer and Mirella Lapata. 2014.
\newblock \href {http://www.aclweb.org/anthology/P14-1068} {Learning grounded
  meaning representations with autoencoders}.
\newblock In \emph{Proceedings of the 52nd Annual Meeting of the Association
  for Computational Linguistics (Volume 1: Long Papers)}, pages 721--732,
  Baltimore, Maryland. Association for Computational Linguistics.

\bibitem[{Socher et~al.(2013)Socher, Chen, Manning, and
  Ng}]{DBLP:conf/nips/SocherCMN13}
Richard Socher, Danqi Chen, Christopher~D. Manning, and Andrew~Y. Ng. 2013.
\newblock \href
  {http://papers.nips.cc/paper/5028-reasoning-with-neural-tensor-networks-for-knowledge-base-completion}
  {Reasoning with neural tensor networks for knowledge base completion}.
\newblock In \emph{Advances in Neural Information Processing Systems 26: 27th
  Annual Conference on Neural Information Processing Systems 2013. Proceedings
  of a meeting held December 5-8, 2013, Lake Tahoe, Nevada, United States.},
  pages 926--934.

\bibitem[{Socher et~al.(2011)Socher, Pennington, Huang, Ng, and
  Manning}]{socher-EtAl:2011:EMNLP}
Richard Socher, Jeffrey Pennington, Eric~H. Huang, Andrew~Y. Ng, and
  Christopher~D. Manning. 2011.
\newblock \href {http://www.aclweb.org/anthology/D11-1014} {Semi-supervised
  recursive autoencoders for predicting sentiment distributions}.
\newblock In \emph{Proceedings of the 2011 Conference on Empirical Methods in
  Natural Language Processing}, pages 151--161, Edinburgh, Scotland, UK.
  Association for Computational Linguistics.

\bibitem[{Tian et~al.(2016)Tian, Okazaki, and
  Inui}]{tian-okazaki-inui:2016:P16-1}
Ran Tian, Naoaki Okazaki, and Kentaro Inui. 2016.
\newblock \href {http://www.aclweb.org/anthology/P16-1121} {Learning
  semantically and additively compositional distributional representations}.
\newblock In \emph{Proceedings of the 54th Annual Meeting of the Association
  for Computational Linguistics (Volume 1: Long Papers)}, pages 1277--1287,
  Berlin, Germany. Association for Computational Linguistics.

\bibitem[{Titov and Khoddam(2015)}]{titov-khoddam:2015:NAACL-HLT}
Ivan Titov and Ehsan Khoddam. 2015.
\newblock \href {http://www.aclweb.org/anthology/N15-1001} {Unsupervised
  induction of semantic roles within a reconstruction-error minimization
  framework}.
\newblock In \emph{Proceedings of the 2015 Conference of the North American
  Chapter of the Association for Computational Linguistics: Human Language
  Technologies}, pages 1--10, Denver, Colorado. Association for Computational
  Linguistics.

\bibitem[{Toutanova and Chen(2015)}]{toutanova-chen:2015:CVSC}
Kristina Toutanova and Danqi Chen. 2015.
\newblock \href {http://www.aclweb.org/anthology/W15-4007} {Observed versus
  latent features for knowledge base and text inference}.
\newblock In \emph{Proceedings of the 3rd Workshop on Continuous Vector Space
  Models and their Compositionality}, pages 57--66, Beijing, China. Association
  for Computational Linguistics.

\bibitem[{Toutanova et~al.(2016)Toutanova, Lin, Yih, Poon, and
  Quirk}]{toutanova-EtAl:2016:P16-1}
Kristina Toutanova, Victoria Lin, Wen-tau Yih, Hoifung Poon, and Chris Quirk.
  2016.
\newblock \href {http://www.aclweb.org/anthology/P16-1136} {Compositional
  learning of embeddings for relation paths in knowledge base and text}.
\newblock In \emph{Proceedings of the 54th Annual Meeting of the Association
  for Computational Linguistics (Volume 1: Long Papers)}, pages 1434--1444,
  Berlin, Germany. Association for Computational Linguistics.

\bibitem[{Trouillon et~al.(2016)Trouillon, Welbl, Riedel, Gaussier, and
  Bouchard}]{DBLP:conf/icml/TrouillonWRGB16}
Th{\'{e}}o Trouillon, Johannes Welbl, Sebastian Riedel, {\'{E}}ric Gaussier,
  and Guillaume Bouchard. 2016.
\newblock \href {https://dl.acm.org/citation.cfm?id=3045609} {Complex
  embeddings for simple link prediction}.
\newblock In \emph{Proceedings of the 33nd International Conference on Machine
  Learning, {ICML} 2016, New York City, NY, USA, June 19-24, 2016}, pages
  2071--2080.

\bibitem[{Wang et~al.(2017)Wang, Mao, Wang, and
  Guo}]{DBLP:journals/tkde/WangMWG17}
Quan Wang, Zhendong Mao, Bin Wang, and Li~Guo. 2017.
\newblock \href {https://doi.org/10.1109/TKDE.2017.2754499} {Knowledge graph
  embedding: {A} survey of approaches and applications}.
\newblock \emph{{IEEE} Trans. Knowl. Data Eng.}, 29(12):2724--2743.

\bibitem[{Wang et~al.(2014{\natexlab{a}})Wang, Zhang, Feng, and
  Chen}]{wang-EtAl:2014:EMNLP20145}
Zhen Wang, Jianwen Zhang, Jianlin Feng, and Zheng Chen. 2014{\natexlab{a}}.
\newblock \href {http://www.aclweb.org/anthology/D14-1167} {Knowledge graph and
  text jointly embedding}.
\newblock In \emph{Proceedings of the 2014 Conference on Empirical Methods in
  Natural Language Processing (EMNLP)}, pages 1591--1601, Doha, Qatar.
  Association for Computational Linguistics.

\bibitem[{Wang et~al.(2014{\natexlab{b}})Wang, Zhang, Feng, and
  Chen}]{DBLP:conf/aaai/WangZFC14}
Zhen Wang, Jianwen Zhang, Jianlin Feng, and Zheng Chen. 2014{\natexlab{b}}.
\newblock \href
  {https://www.aaai.org/ocs/index.php/AAAI/AAAI14/paper/viewPaper/8531}
  {Knowledge graph embedding by translating on hyperplanes}.
\newblock In \emph{Proceedings of the Twenty-Eighth {AAAI} Conference on
  Artificial Intelligence, July 27 -31, 2014, Qu{\'{e}}bec City, Qu{\'{e}}bec,
  Canada.}, pages 1112--1119.

\bibitem[{Xiao et~al.(2016)Xiao, Huang, and Zhu}]{DBLP:conf/ijcai/xiao16}
Han Xiao, Minlie Huang, and Xiaoyan Zhu. 2016.
\newblock \href {http://www.ijcai.org/Abstract/16/190} {From one point to a
  manifold: Knowledge graph embedding for precise link prediction}.
\newblock In \emph{Proceedings of the Twenty-Fifth International Joint
  Conference on Artificial Intelligence, {IJCAI} 2016, New York, NY, USA, 9-15
  July 2016}, pages 1315--1321.

\bibitem[{Xie et~al.(2017)Xie, Ma, Dai, and Hovy}]{xie-EtAl:2017:Long}
Qizhe Xie, Xuezhe Ma, Zihang Dai, and Eduard Hovy. 2017.
\newblock \href {http://aclweb.org/anthology/P17-1088} {An interpretable
  knowledge transfer model for knowledge base completion}.
\newblock In \emph{Proceedings of the 55th Annual Meeting of the Association
  for Computational Linguistics (Volume 1: Long Papers)}, pages 950--962,
  Vancouver, Canada. Association for Computational Linguistics.

\bibitem[{Yang et~al.(2015)Yang, Yih, He, Gao, and Deng}]{Yang2015}
Bishan Yang, Wen-tau Yih, Xiaodong He, Jianfeng Gao, and Li~Deng. 2015.
\newblock \href {http://arxiv.org/abs/1412.6575} {{Embedding Entities and
  Relations for Learning and Inference in Knowledge Bases}}.
\newblock In \emph{Proceedings of the 3rd International Conference on Learning
  Representations}, pages 1--12.

\end{thebibliography}

\appendix

\section{Out-of-vocabulary Entities in KBC}

Occasionally, a KBC test set may contain entities that never appear in the training data. Such 
out-of-vocabulary (OOV) entities pose a challenge to KBC systems; while some systems address 
this issue by explicitly learn an OOV entity vector \citep{dettmers2018conve}, our approach is described below. For an incomplete triple $\mtriple{h}{r}{?}$ in the test, if $h$ is OOV, 
we replace it with the most frequent entity that has ever appeared as a head of relation $r$ 
in the training data. If the gold tail entity is OOV, we use the zero vector for computing 
the score and the rank of the gold entity. 

Usually, OOV entities are rare and negligible in evaluation; except for the WN18RR test data 
which contains about 6.7\% triples with OOV entities. Here, we also report adjusted scores on WN18RR 
in the setting that all triples with OOV entities are removed from the test. The 
results are shown in Table~\ref{tab:wn18rr-remove-oov}.

\begin{table}[ht]
\centering
\small
\setlength{\tabcolsep}{13pt}
\begin{tabular}{lccc}
\toprule
Model & MR & MRR & H10 \\
\midrule
\textsc{joint} & \textbf{3317} & \textbf{.493} & \textbf{57.2} \\
\textsc{base} & 3435 & .492 & 56.7 \\
\midrule
\textsc{joint+comp} & \textbf{1507} & \textbf{.367} & \textbf{58.7} \\
\textsc{base+comp} & 1629 & .332 & 58.0 \\
\bottomrule
\end{tabular}
\caption{Adjusted scores on WN18RR.}
\label{tab:wn18rr-remove-oov}
\end{table}

% there are entities 

% WN18RR contains 212 out-of-vocabulary (OOV) entities
% (i.e., they do not appear in the training set) in the test set.
% Since many test triples (about 6.7\% out of all the test triples) has an OOV entity in their head or tail
% and can affect the overall performance,
% we use the following procedure while testing such triples.
% When we predict $h$ given $\mtriple{?}{r}{t}$ s.t. $h$ is an OOV entity,
% we assume $\vec{v}_h$ is a zero vector and the rest of procedure is identical to the above.
% When we predict $h$ given $\mtriple{?}{r}{t}$ s.t. $t$ is an OOV entity,
% we replace
% $t$ with
% pseudo $t'$ that most frequently occurs
% in the training set in the form $\mtriple{?}{r}{t'}$,
% and the rest is identical to the above.
% FB15k-237 also has a small number of triples that contain OOV entities in the test set,
% but for simplicity we omit such triples from the test set.

% \section{Initialization}

\end{document}